\PassOptionsToPackage{hyphens}{url}

\documentclass[10pt,twocolumn,letterpaper,]{article}
\usepackage[colorlinks, linkcolor=darkred, urlcolor=darkred, citecolor=teal]{hyperref}
\usepackage{cvpr}              
\usepackage{tcolorbox}

\usepackage[utf8]{inputenc}
\usepackage[T1]{fontenc}
\usepackage{booktabs}       
\usepackage{amsfonts}       
\usepackage{nicefrac}       
\usepackage{microtype}      
\usepackage{amsmath, amssymb, amsthm}
\usepackage{colortbl}
\usepackage{pifont} 
\usepackage{multirow}
\usepackage{multicol}
\usepackage{makecell}
\usepackage{graphicx}
\usepackage{diagbox}
\usepackage{subcaption}
\usepackage[utf8]{inputenc}
\usepackage{svg}
\usepackage{comment} 

\usepackage{float}
\usepackage{enumitem}
\usepackage{caption}
\usepackage{graphicx}
\usepackage{caption}
\usepackage{booktabs}
\usepackage{graphicx}    
\usepackage{caption}     
\usepackage{subcaption}  
\usepackage{newunicodechar}
\newunicodechar{，}{,}
\usepackage[table,dvipsnames]{xcolor}
\usepackage{graphicx}
\usepackage{algorithm}

\usepackage{algpseudocode}
\usepackage{wrapfig}
\usepackage{enumitem}
\usepackage{adjustbox}
\setenumerate[1]{itemsep=0pt,partopsep=0pt,parsep=\parskip,topsep=5pt}
\setitemize[1]{itemsep=0pt,partopsep=0pt,parsep=\parskip,topsep=5pt}
\setdescription{itemsep=0pt,partopsep=0pt,parsep=\parskip,topsep=5pt}
\algrenewcommand\algorithmiccomment[1]{\hfill\textit{\color{gray}\# #1}}
\definecolor{darkblue}{rgb}{0, 0.12, 0.55}
\definecolor{darkgreen}{rgb}{0, 0.55, 0.12}
\definecolor{darkred}{rgb}{0.6,0,0}
\definecolor{darkgreen}{rgb}{0,0.6,0}
\newcommand{\rone}[1]{{\color{ForestGreen} #1}}
\newcommand{\rthree}[1]{{\color{Magenta} #1}}
\newcommand{\rfive}[1]{{\color{RoyalBlue} #1}}
\usepackage{nicematrix}
\usepackage{booktabs}
\usepackage{lineno}

\definecolor{MyTeal}{HTML}{206070}

\newcommand{\algstep}[1]{%
    \Statex\hspace{-\algorithmicindent}\textbf{\quad  #1}%
}

\newcommand{\LineComment}[1]{\Statex \textit{\color{gray}\qquad\# #1}}

\newcommand{\name}{Hunyuan-GameCraft-2: Instruction-following Interactive Game World Model}

\title{\name}

\author{
  Tencent Hunyuan\thanks{ corresponding author (Email: \url{qinglinlu@tencent.com})}
}

\begin{document}

\author{Junshu Tang\textsuperscript{1}\footnotemark[1] \quad  Jiacheng Liu\textsuperscript{1,2}\footnotemark[1] \footnotemark[2] \quad Jiaqi Li\textsuperscript{1}\footnotemark[1] \footnotemark[2] \quad Longhuang Wu\textsuperscript{1} \quad Haoyu Yang\textsuperscript{1} \\  Penghao Zhao\textsuperscript{1} \quad Siruis Gong\textsuperscript{1} \quad Xiang Yuan\textsuperscript{1} \quad Shuai Shao\textsuperscript{1} \quad Linfeng Zhang\textsuperscript{2} \quad Qinglin Lu\textsuperscript{1}\footnotemark[3]\\
\\
\textsuperscript{1} Tencent Hunyuan \quad \textsuperscript{2} Shanghai Jiao Tong University  \\
\url{https://hunyuan-gamecraft-2.github.io/}
}

\twocolumn[{
\renewcommand\twocolumn[1][]{#1}
\maketitle
\centering
\vspace{-7mm}
\begin{adjustbox}{width=\linewidth, center}
    \includegraphics{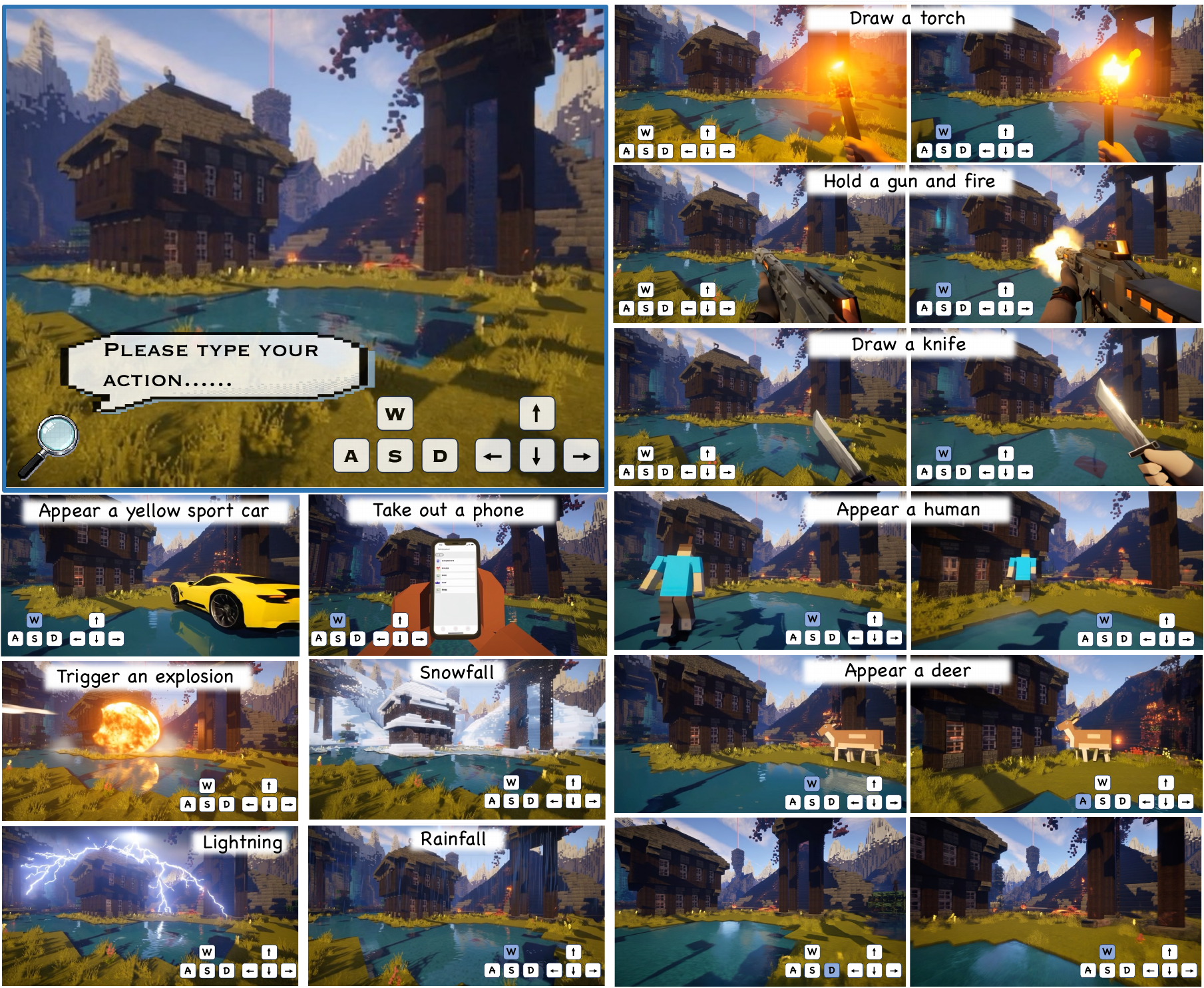} 
\end{adjustbox}
 \captionsetup{type=figure}
\vspace{-3mm}
\caption{\emph{\textbf{Hunyuan-GameCraft-2}} advances generative game world models from static game scene video synthesis to open-ended, instruction-following interactive simulation.
We simulate a series of action signals from a single image. The left and right frames depict key moments from game video sequences generated in response to different action inputs. Our model can accurately produce content aligned with each interaction, supports high-fidelity game video generation with temporal and 3D consistency. As shown, instructions such as \textit{``draw a torch''} or \textit{``hold a gun and fire''}, and key/mouse action dynamically guide camera motion and dynamic video content editing, producing temporally coherent, causally grounded interactive videos with realistic scene continuity and a consistent style.
In this case, W, A, S, D represent transition movement and ↑, ←, ↓, → denote changes in view angles, and white denotes an idle state.
}
\vspace{-3mm}
\label{fig:teaser}
}]

\renewcommand{\thefootnote}{\fnsymbol{footnote}}
\footnotetext[1]{Equal Contribution.}
\footnotetext[2]{Work is done during the internship at Tencent Hunyuan.}
\footnotetext[3]{Corresponding author.}

\newpage
\newpage
\begin{abstract}

Recent advances in generative world models have enabled remarkable progress in creating open-ended game environments, evolving from static scene synthesis toward dynamic, interactive simulation. However, current approaches remain limited by rigid action schemas and high annotation costs, restricting their ability to model diverse in-game interactions and player-driven dynamics.
To address these challenges, we introduce \textbf{\textit{Hunyuan-GameCraft-2}}, a new paradigm of instruction-driven interaction for generative game world modeling. Instead of relying on fixed keyboard inputs, our model allows users to control game video contents through natural language prompts, keyboard, or mouse signals, enabling flexible and semantically rich interaction within generated worlds.
We formally define the concept of Interactive Video Data and develop an automated pipeline that converts large-scale, unstructured text–video pairs into causally aligned interactive datasets. Built upon a 14B image-to-video Mixture-of-Experts (MoE) foundation model, our model incorporates a text-driven interaction injection mechanism for fine-grained control over camera motion, character behavior, and environment dynamics. We introduce an interaction-focused benchmark, InterBench to evaluate interaction performance comprehensively.
Extensive experiments demonstrate that our model generates temporally coherent, and causally grounded interactive game videos that faithfully respond to diverse and free-form user instructions such as ``open the door'', ``draw a torch'', or ``trigger an explosion''.

\end{abstract}
    
\section{Introduction}
\label{sec:intro}

The rapid advancement of diffusion models~\cite{StableDiffusion,DiT,DM,wan_wan_2025,kong2024hunyuanvideo,li_hunyuan-dit_2024} has significantly advanced dynamic game content creation~\cite{Hunyuan-GameCraft,zhang2025matrixgame,he2025matrix}. 
Beyond static image or short video synthesis, recent cutting-edge achievements, from RTFM~\cite{rtfm} and to Genie series~\cite{genie3}, mark world model can serve as the foundation for immersive, controllable virtual experiences, marking a crucial step toward AI-driven \textbf{``playable world''} that can both simulate and respond to user intent.

Existing world models can be categorized into \textit{\textbf{3D-based}} and \textit{\textbf{video-based}} approaches. 3D-based world models~\cite{worldlabs2024,rtfm,team_hunyuanworld_2025,liu2025worldmirror,li_flashworld_2025,huang2025voyager} emphasize geometric consistency and physical accuracy, enabling detailed world reconstruction and memory persistence. However, they are often limited to scripted or static interactions, lacking the creative flexibility and open-ended gameplay dynamics essential for interactive game environments.
With recent improvements in video foundation models~\cite{wan_wan_2025,kong2024hunyuanvideo,veo3,openai2024sora}, the video-based technical pathway~\cite{genie3,parkerholder2024genie2,feng2024matrix,zhang2025matrixgame,he2025matrix,xiao_worldmem_2025,yu2025gamefactory,Hunyuan-GameCraft} has shown remarkable potential. 
These works learn world dynamics directly from large-scale video data~\cite{ real10k,Sekai,che2024gamegen,feng2024matrix} through implicit end-to-end representation learning. 
Notably, the Genie series~\cite{parkerholder2024genie2,genie3} introduces latent action modeling to simulate player-driven physical interactions, while Matrix-Game~\cite{zhang2025matrixgame} and Hunyuan-GameCraft~\cite{Hunyuan-GameCraft} integrate discrete gameplay actions (e.g., W/A/S/D, mouse movements) into a unified representation space, achieving continuous, high-fidelity video generation that responds to user inputs.

These frontier works mark a fundamental shift in focus from the world's static appearance  ``\textit{what the world \textbf{looks like}}'' to its interactive dynamics ''\textit{how we \textbf{interact with} it}''. Consequently, compelling us to  rigorously define the concept of ``\textbf{\textit{interaction}}'' within the context of world models, especially in game scenarios.

We formally define interaction in world models as ''\textit{actions executed by an explicit agent that trigger state transitions in the environment with clear causal relationships and physical or logical validity.}'' 
This definition encompasses diverse input modalities, from mouse and keyboard operations~\cite{Hunyuan-GameCraft,yu2025gamefactory,xiao_worldmem_2025,feng2024matrix,zhang2025matrixgame} to embodied motion sensing~\cite{long2025surveylearningembodiedintelligence}. Grounded in this perspective, 
two key challenges hinder this progress: (1) the lack 
of a formal definition and a scalable construction pipeline for 
interactive video data, 
and (2) multi-turn interactions in long video generation while maintaining video quality and interaction accuracy.

To address these challenges, we present \textbf{Hunyuan-GameCraft-2}, an interactive game world model for free-form instruction-following control. We begin by formally defining interaction within the context of generative world models, and develop two automated pipelines for interactive video data construction and refinement.
These pipelines, for the first time, enable the efficient transformation of large-scale, unstructured text–video pairs into open-domain interactive datasets enriched with implicit causal labels.

For model training, our model integrates text-based instructions and keyboard/mouse action signals into a unified controllable video generator, enabling flexible, semantically grounded, and causally consistent interaction within dynamic game environments. To support efficient long-horizon video generation, we employ a comprehensive autoregressive distillation strategy that transfers the bidirectional video generator into a causal autoregressive model.
Subsequently, a randomized image-to-long-video extension tuning scheme is introduced to alleviate error accumulation during extended rollouts, ensuring stable and coherent long-form generation.
For multi-turn interactive inference, we following LongLive~\cite{yang2025longliverealtimeinteractivelong} to employ a KV-recache mechanism to
enhance the accuracy and stability of multi-turn interactions
in autoregressive long video generation.
In addition, we incorporate several engineering acceleration optimizations, boosting the model’s inference speed to 16 FPS, enabling real-time interactive video generation.

To comprehensively evaluate interactive performance across different models, we introduce InterBench, a new benchmark that systematically measures key dimensions of interactive behavior — including interaction completeness, action effectiveness, causal coherence, and physical plausibility.
Extensive experiments on InterBench and general video-quality metrics demonstrate the effectiveness of our framework, achieving state-of-the-art performance in generating interactive videos that faithfully respond to user instructions while maintaining high visual fidelity and temporal coherence.

In general, our main contributions are as follows:

\begin{itemize}
\item We propose a unified controllable video generation framework integrating text, keyboard, and mouse signals for semantically grounded interactions.

\item We leverage autoregressive distillation and randomized long-video tuning to ensure efficient and stable long-horizon generation, with KV-recache for multi-turn inference and real-time 16 FPS performance through engineering optimizations.

\item Through extensive quantitative and qualitative experiments, we comprehensively validate the effectiveness of our proposed framework, demonstrating superior performance in generating interactive videos that faithfully respond to user instructions while maintaining visual quality and temporal coherence.
\end{itemize}
\section{Related Works}
\label{sec:related_works}

\subsection{Long Video Extension}

Maintaining temporal coherence in long video generation is a principal challenge, primarily stemming from the "train-short-test-long" discrepancy in diffusion models, which often causes semantic drift and accumulating artifacts. To surmount this, one major line of work seeks to better align the training process with inference conditions. Methods such as Self-Forcing~\citep{huang2025selfforcingbridgingtraintest} condition the model on its own predictions to simulate error accumulation, while Rolling-Forcing~\citep{liu2025rollingforcingautoregressivelong} incrementally updates context through rolling windows. A complementary strategy integrates explicit memory structures, as seen in Memory-Forcing~\citep{huang2025memoryforcingspatiotemporalmemory} and StreamingT2V~\citep{henschel2024streamingt2v}, to preserve long-range dependencies and global dynamics. Beyond adapting the existing training loop, other research explores more fundamental shifts in the generative paradigm. These include alternative formulations like next-frame prediction models~\citep{gao2024vid,gu2025long}, hybrid diffusion-autoregressive frameworks such as DiffusionForcing~\citep{chen2024diffusion}, and test-time adaptation for inference refinement~\citep{dalal2025one}. The research frontier is also advancing toward interactive and structured synthesis. For example, LongLive~\citep{yang2025longliverealtimeinteractivelong} introduces a KV-recache mechanism for responsive semantic control, and MAGI-1~\citep{ai2025magi1autoregressivevideogeneration} autoregressively generates temporal blocks to mitigate error propagation through explicit partitioning.

\subsection{Interactive Video-based World Model}
Unlike traditional video generation models which produce predetermined sequences, interactive world models dynamically respond to user inputs, enabling the creation of explorable and playable game environments(Tab.~\ref{tab:compare_works}).

Early explorations in this domain often utilized game environments like Minecraft as a testbed. Models such as MineWorld~\cite{guo2025mineworldrealtimeopensourceinteractive}, Matrix~\cite{feng2024matrix}, and GameFactory~\cite{yu2025gamefactory} demonstrated the ability to generate video conditioned on discrete user actions, typically keyboard and mouse inputs. Similarly, Yume~\cite{mao2025yumeinteractiveworldgeneration} generates interactive video from a single image prompt controlled by discrete keyboard commands. While pioneering, these models were often limited to specific games and simple action spaces. Nevertheless, the scope of interaction these models support remains highly limited.

Subsequent research advanced generalization and long-term consistency. Genie2~\cite{parkerholder2024genie2} introduced a foundation model capable of generating diverse, action-controllable 2D worlds from single images. To tackle the challenge of consistency in extended simulations, WorldMem~\cite{xiao_worldmem_2025} introduced a memory bank framework to address long-term consistency issues, while recent works like PAN~\cite{panteam2025panworldmodelgeneral} also focus on achieving interactive long-range world simulation. 

Building upon these foundations, researchers began to explore more flexible interaction. GameGen-X~\cite{che2024gamegen} integrates multi-modal control signals for open-world games. Critically, Genie3~\cite{genie3} and Hunyuan-GameCraft~\cite{Hunyuan-GameCraft} advance this paradigm by unifying discrete keyboard and mouse signals into a shared, continuous action space. This emerging fusion of direct controls and language prompts shows immense potential. However, prompts in these latest works are predominantly used for world setup and high-level guidance, rather than as a direct, interactive control mechanism. Consequently, the richness of interaction is still fundamentally constrained by the discrete nature of physical input devices.

\begin{table*}[!hbt]
\centering
\renewcommand{\arraystretch}{1.35}
\newcommand{\gou}{\textcolor{ForestGreen}{\ding{52}}} 
\newcommand{\cha}{\textcolor{Red}{\ding{55}}}          
\newcommand{\na}{---}                                   

\caption{Comparison of representative interactive world models across data, action space, generalization, and runtime properties.}

\label{tab:compare_works}

\resizebox{\textwidth}{!}{
\begin{tabular}{l|cccccccc}
\toprule

\textbf{Model} & \textbf{Resolusions} & \textbf{Training Data} & \textbf{Action type} & \textbf{Action space} & \textbf{Action Generalizable} & \textbf{Scene dynamic} & \textbf{Scene memory} & \textbf{Real time} \\
\midrule

\textbf{GameNGen}~\cite{valevski2024diffusion} & 240p & Gameplay& Keyboard & Discrete & Closed & \cha & \gou & \cha \\
\textbf{Oasis}~\cite{oasis} & 360p & Gameplay video & Key+Mouse & Discrete & Closed & \cha & \cha & \gou \\
\textbf{GameGen-X}~\cite{che2024gamegen} & 720p & Gameplay video & Key+Mouse & Discrete & Closed & \cha & \gou & \cha \\
\textbf{Matrix}~\cite{feng2024matrix} & 720p & Gameplay + Rendered & Key & Discrete & Closed & \gou & \cha & \gou \\
\textbf{Matrix-Game}~\cite{zhang2025matrixgame} & 720p & Gameplay + Rendered & Key+Mouse & Discrete & Closed & \gou & \cha & \cha \\
\textbf{Genie 2}~\cite{parkerholder2024genie2} & 720p & Unknown & Key+Mouse & Unknown & Closed & \gou & \cha & \cha \\

\textbf{GameFactory}~\cite{yu2025gamefactory} & 360p & Gameplay video & Key+Mouse & Discrete & Closed & \gou & \cha & \cha \\

\textbf{GameCraft}~\cite{Hunyuan-GameCraft} & 720p &Gameplay + Rendered & Key+Mouse & Continuous & Closed & \gou & \gou & \cha \\
\textbf{Genie 3}~\cite{genie3} & 720p & Unknown & Key+Mouse & Unknown & Unknown & \gou & \gou & \gou \\
\midrule
\rowcolor{gray!15}
\textbf{GameCraft-2} &
480p &
\makecell{Gameplay + Rendered \\+ Interaction Data} &
\makecell{Key+Mouse \\+ \textbf{Prompt-based Instruction}} &
Continuous &
\textbf{Open-ended} &
\gou & \gou & \gou \\
\bottomrule
\end{tabular}}
\end{table*}


\subsection{Text-guided Video Generation and Editing}

Text-based control over video synthesis has advanced significantly through two main paradigms: enhancing semantic understanding and executing structured plans.

The first paradigm focuses on \textit{enriching the initial prompt}. This is achieved by fusing representations from Large Language Models (LLMs) for more nuanced inputs~\citep{tan2024mimirimprovingvideodiffusion, liu2024llm4genleveragingsemanticrepresentation}, using LLMs to rephrase or expand simple queries~\citep{zhang2025riset2vrephrasinginjectingsemantics}, or employing lightweight adapters to bridge domain gaps~\citep{zhao2024bridgingdifferentlanguagemodels, zhang2025riset2vrephrasinginjectingsemantics}. The second, more sophisticated paradigm treats text as a script or plan to be executed. Here, LLMs act as   \textit{``directors''}, decomposing high-level prompts into a sequence of frame-by-frame descriptions for temporally evolving scenes~\citep{huang2023freebloomzeroshottexttovideogenerator, hong2024direct2vlargelanguagemodels}. This concept extends to orchestrating complex, multi-scene videos with explicit spatial layouts and consistency constraints~\citep{lian2024llmgroundeddiffusionenhancingprompt, lin2024videodirectorgptconsistentmultiscenevideo}. A related approach is found in video editing, where textual instructions guide discrete tasks like style transfer or object manipulation, often within video-to-video frameworks that enable zero-shot or end-to-end control~\citep{li2023vidtomevideotokenmerging, cheng2023consistentvideotovideotransferusing, qin2024instructvid2vidcontrollablevideoediting, qi2023fatezerofusingattentionszeroshot, khachatryan2023text2videozerotexttoimagediffusionmodels}.

Despite their power, these methods are fundamentally non-interactive. Whether enhancing a prompt or executing a script, they perform one-off transformations based on a static, predefined set of commands. They lack the core concepts of state transition and continuous feedback, where an action perpetually redefines future possibilities. In stark contrast, Hunyuan-GameCraft-2 introduces true interaction, where user prompts continuously drive the evolution of a dynamic world state, addressing a fundamentally different objective than either planned generation or scripted editing.


\section{Interactive Video Data Construction}
\label{sec:data}

Current video data suitable for training interactive world models remain scarce. Real-world captured videos offer high realism but are costly, time-consuming to collect, and difficult to scale. Simulation-based generation using engines such as \textit{Unreal Engine} provides strong controllability over viewpoint and interactions, yet the heavy modeling and rendering costs restrict scene diversity. Internet videos from platforms like \textit{YouTube} offer massive volume and variety, but their highly inconsistent quality and abundant noise demand complex cleaning pipelines. Public academic datasets, while well-annotated and reliable, are limited in scale and domain coverage. As a result, none of these sources simultaneously satisfy the requirements of interactivity, large scale, and broad diversity, leaving high-quality interactive video data fundamentally insufficient. This scarcity highlights the need for a clearer understanding of what truly qualifies as \textit{interactive} data, which we formalize in the following analysis.

\subsection{Definition of  Interactive Video Data}

\begin{tcolorbox}[
title=\textbf{
}, coltitle=black, colframe=gray!40]
Interactive Video Data refers to a temporal sequence that explicitly records a 
\rone{\textbf{causally driven state-transition process}}, in which agents or the environment 
transition from a clearly defined \rthree{\emph{\textbf{initial state}}} to a significantly different 
\rfive{\emph{\textbf{final state}}}. The importance of such data lies in its ability to 
faithfully capture \underline{\emph{how an event evolves over time}}, rather than in visual complexity.
\end{tcolorbox}

\begin{figure*}[t!]
    \centering
    \includegraphics[width=\linewidth]{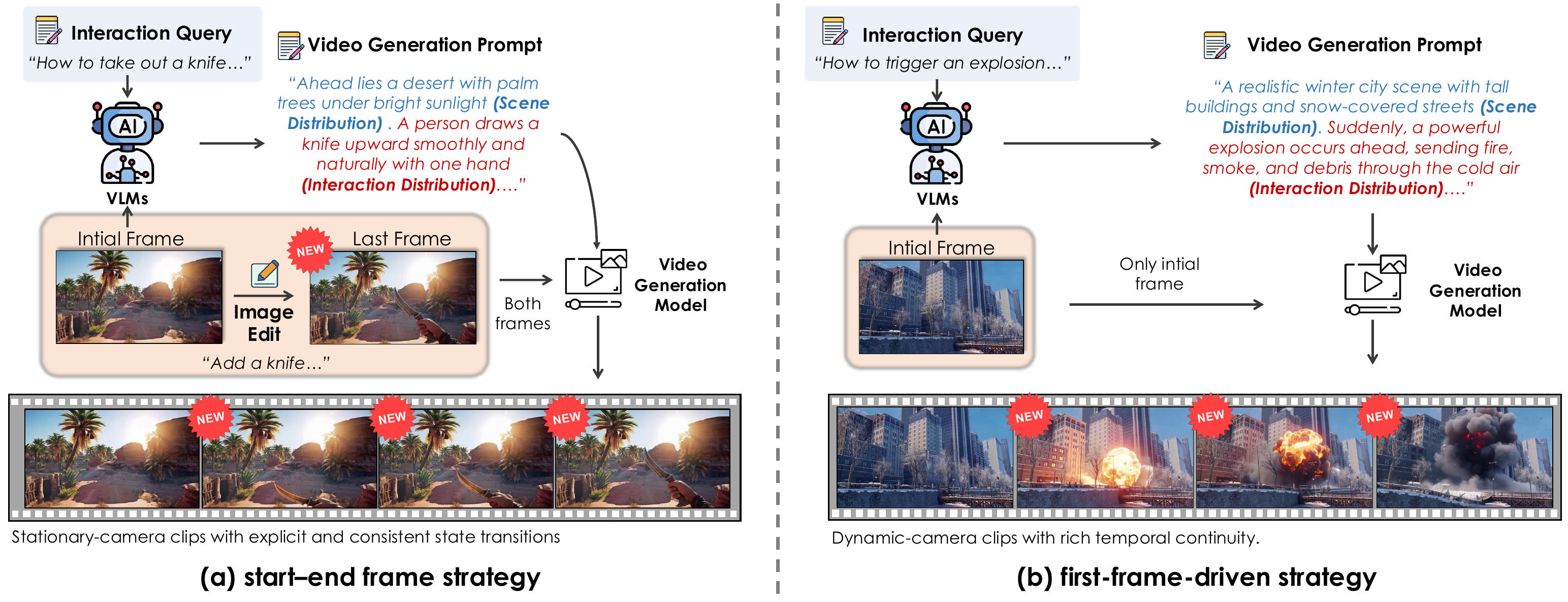}
    \vspace{-6mm}
    \caption{\textbf{Showcase of our Synthetic Interaction Video Pipeline.}
\textbf{\textit{(a) The start–end frame strategy}} uses a VLM and an image-editing model to construct both initial and edited target frames, enabling controlled state transitions for stationary-camera scenarios.
\textbf{\textit{(b) The first-frame-driven strategy}} relies solely on the initial frame and VLM-generated prompts, allowing the video generator to create dynamic, motion-rich interactions with flexible camera movement.}
    \label{fig:construction_pipeline}
    \vspace{-3mm}
\end{figure*}

A video segment is considered \textit{interactive} if it satisfies any of the following properties:

\begin{itemize}
    \item \textbf{Significant State Transition.} 
    The video must contain a recognizable and non-trivial macroscopic change of state. 
    It should present clearly distinguishable \emph{pre-condition} and \emph{post-condition} states, 
    with the temporal content between them forming the \emph{transition process}.

    \item \textbf{Subject Emergence or Interaction.} 
    The main content involves explicit subjects, including:
    \begin{enumerate}[leftmargin=26pt]
        \item \emph{Emergence}: a new subject appears in a previously empty context.
        \item \emph{Action-driven}: a subject performs an action that changes its own state 
        or affects the environment.
    \end{enumerate}

    \item \textbf{Scene Shift or Evolution.} 
    The video records a fundamental shift or evolution of the scene or background, 
    rather than minor or random perturbations.
\end{itemize}

Interactive videos thus possess \textbf{explicit causal structure}, \textbf{clear state transitions}, and \textbf{perceivable action agents}, enabling world models to learn interpretable action–outcome mappings.  Following this definition, we systematically organize interactive data into three principal categories to structure our analysis: \textbf{(1) Environmental Interactions}, which encompass global or local scene changes; \textbf{(2) Actor Actions}, which are driven by an embodied agent; and \textbf{(3) Entity and Object Appearances}, which involve the introduction of new subjects. To facilitate a nuanced evaluation, each category is further divided into \textit{simple} and \textit{complex} settings, reflecting varying degrees of difficulty. Specific examples for each category are provided in Appendix~\ref{sec:appendix_categories}.

\begin{figure*}[hbt!]
    \centering
    \includegraphics[width=0.85\linewidth]{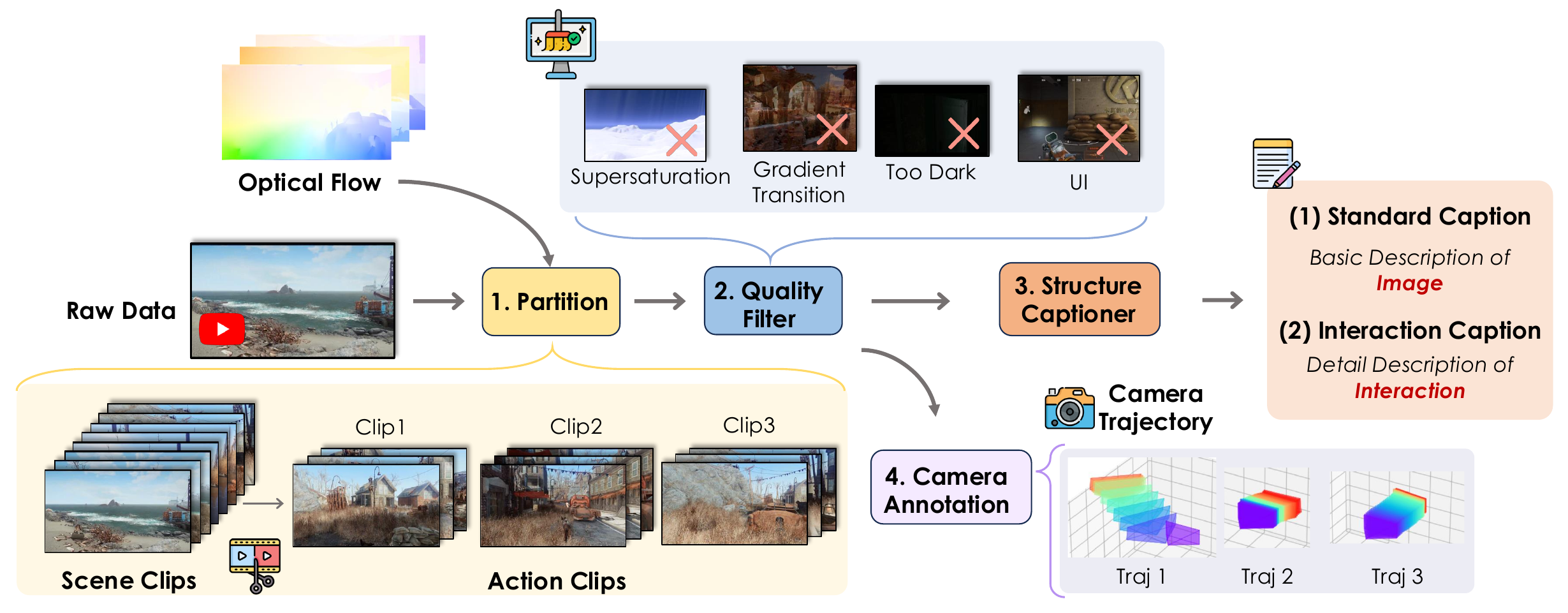}
\caption{\textbf{Pipeline of the Data Curation System.}
Our pipeline consists of four stages: 
\textbf{\textit{(1) Partition,}} which segments long gameplay videos into scene- and action-level clips using scene detection and optical-flow cues;
\textbf{\textit{(2) Quality Filtering}}, which removes low-quality frames via visual assessment, luminance checks, and VLM-based semantic filtering;
\textbf{\textit{(3) Structured Captioning}}, which produces both standard and interaction-centric captions for each clip;
\textbf{\textit{(4) Camera Annotation}}, which reconstructs 6-DoF trajectories to capture viewpoint motion. 
These steps convert raw gameplay footage into clean, structured, and interaction-aware training data.}
    \label{fig:curation_pipeline}
    \vspace{-3mm}
\end{figure*}

\subsection{Synthetic Data Construction}

To address the scarcity and high annotation cost of interactive video data, we propose a controllable Synthetic Interaction Video Pipeline for large-scale automated production. While generating synthetic data for training video models has been underexplored, we argue it is now feasible by leveraging the advanced world knowledge and visual representation capabilities of recent foundation models. The effectiveness of our pipeline in producing diverse, high-quality data is showcased in Appendix~\ref{sec:appendix_dataset_showcase} (Figs.~\ref{fig:Construction_data_enirvonment}-\ref{fig:Construction_data_entity_appearance}).

We generate interactive videos starting from an initial frame $F_t$. To handle diverse visual contexts, we first employ a Vision-Language Model (VLM) to analyze $F_t$ and, guided by a high-level instruction (e.g., \textit{``taking out a torch''}), generate a customized, scene-specific prompt. Based on the interaction type, we then apply one of two distinct strategies:
\begin{enumerate}
    \item \textbf{Start-End Frame Strategy:} For stationary scenes requiring explicit state transitions (e.g., environmental changes like \textit{``making it snow''}), a VLM guides an image editing model to generate a target end-frame $F'_t$. This provides strong controllability over the final state.
    \item \textbf{First-Frame-Driven Strategy:} For dynamic actions involving significant camera motion (e.g., \textit{``opening a door''}), the model generates freely from only the initial frame. This approach avoids distortions and yields smoother camera movement and temporal continuity.
\end{enumerate}

Sourcing specific initial frames for certain interactions, such as \textit{``opening a door''}, is a significant bottleneck, as manual curation is both costly and inefficient. To address this, we leverage an advanced text-to-image model (e.g., HunyuanImage-3.0~\cite{cao2025hunyuanimage}), to synthesize these requisite frames on demand, providing a scalable source of high-quality inputs for our video generation pipeline.

\begin{figure*}[hbt!]
    \centering
    \includegraphics[width=0.9\linewidth]{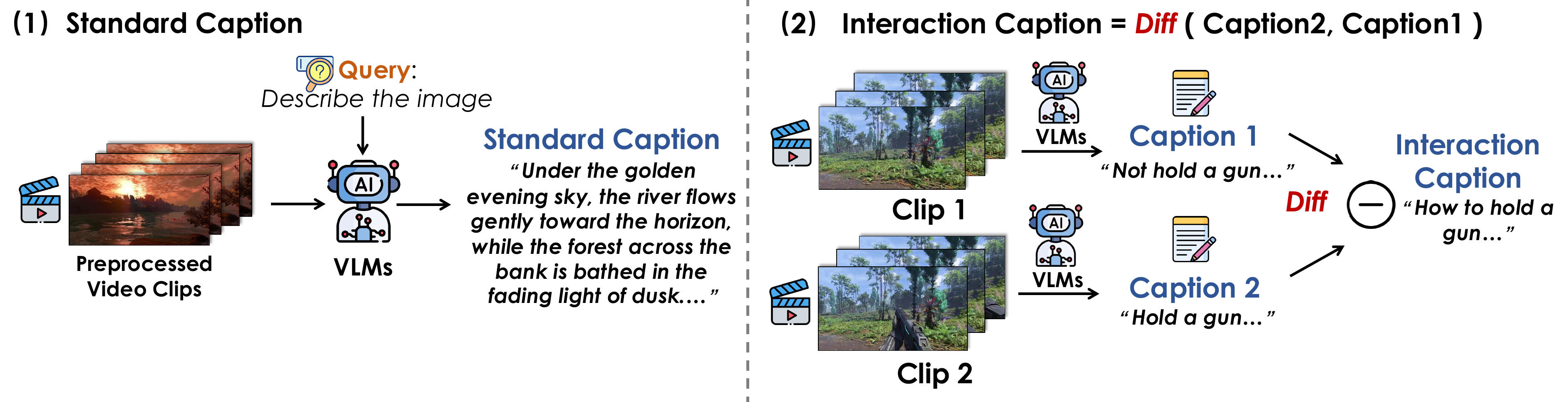}
    \vspace{-4mm}
\caption{\textbf{Pipeline of the Caption Generation System.}
The system produces two forms of captions: a \textit{\textbf{standard caption}} that describes the visual content of each clip, and an \textit{\textbf{interaction caption}} derived by computing the semantic difference between consecutive clips. This enables both scene-level descriptions and explicit interaction-oriented annotations for supervision.}
    \label{fig:caption_pipeline}
    \vspace{-4mm}
\end{figure*}


\subsection{Game Scene Data Curation}

We build our dataset from over 150 AAA games (e.g., \textit{Assassin’s Creed}, \textit{Cyberpunk 2077}), which provides extensive diversity in environments, lighting, artistic styles, and camera viewpoints is showcased in Appendix~\ref{sec:appendix_dataset_showcase} Figs.~\ref{fig:curation_data_showcase} and~\ref{fig:curation_data_showcase_2}.

\noindent \textbf{Scene and Action-aware Data Partition.}
We employ a two-stage partitioning strategy to process the raw videos. First, PySceneDetect~\cite{Castellano_PySceneDetect} segments long videos into visually coherent 6-second clips. Subsequently, we use RAFT-based optical flow~\cite{teed2020raft} to localize fine-grained action boundaries, ensuring each clip preserves temporal integrity for training.

\noindent \textbf{Data Filtering.}
To ensure data quality, we perform a three-stage filtering process. A learning-based model first removes low-fidelity or artifact-heavy frames~\cite{kolors}. Next, luminance filtering eliminates scenes that are poorly lit~\cite{bradski2000opencv}. Finally, a VLM-based semantic check verifies content consistency across frames, retaining only clips with clean visual structure and accurate motion alignment~\cite{wang2024qwen2}.

\noindent \textbf{Camera Annotation.}
We reconstruct 6-DoF camera trajectories for each clip using VIPE~\cite{huang2025vipe}. This process yields frame-by-frame translational and rotational motion estimates, providing precise metadata for training camera-aware models and enforcing spatio-temporal consistency.

\noindent \textbf{Structured Captioning.}
To provide interaction-aware supervision, we devise a structured captioning scheme with two components. First, a \textbf{Standard Caption} ($C_t$), generated by a VLM for each clip, describes the static visual content. Second, an \textbf{Interaction Caption} ($I_{t \rightarrow t+1}$) captures the state transition between adjacent clips. This interaction is computed as the semantic difference between their respective standard captions:
\[
I_{t \rightarrow t+1} = \Delta(\Phi(C_{t+1}), \Phi(C_{t})),
\]
where $\Phi$ is a semantic encoder and $\Delta$ is a difference operator. This dual-component approach enables the model to jointly learn \textbf{appearance-level perception} (from $C_t$) and \textbf{action-level reasoning} (from $I_{t \rightarrow t+1}$).

\section{Method}
\label{sec:method}

\begin{figure*}
    \centering
    \includegraphics[width=\linewidth]{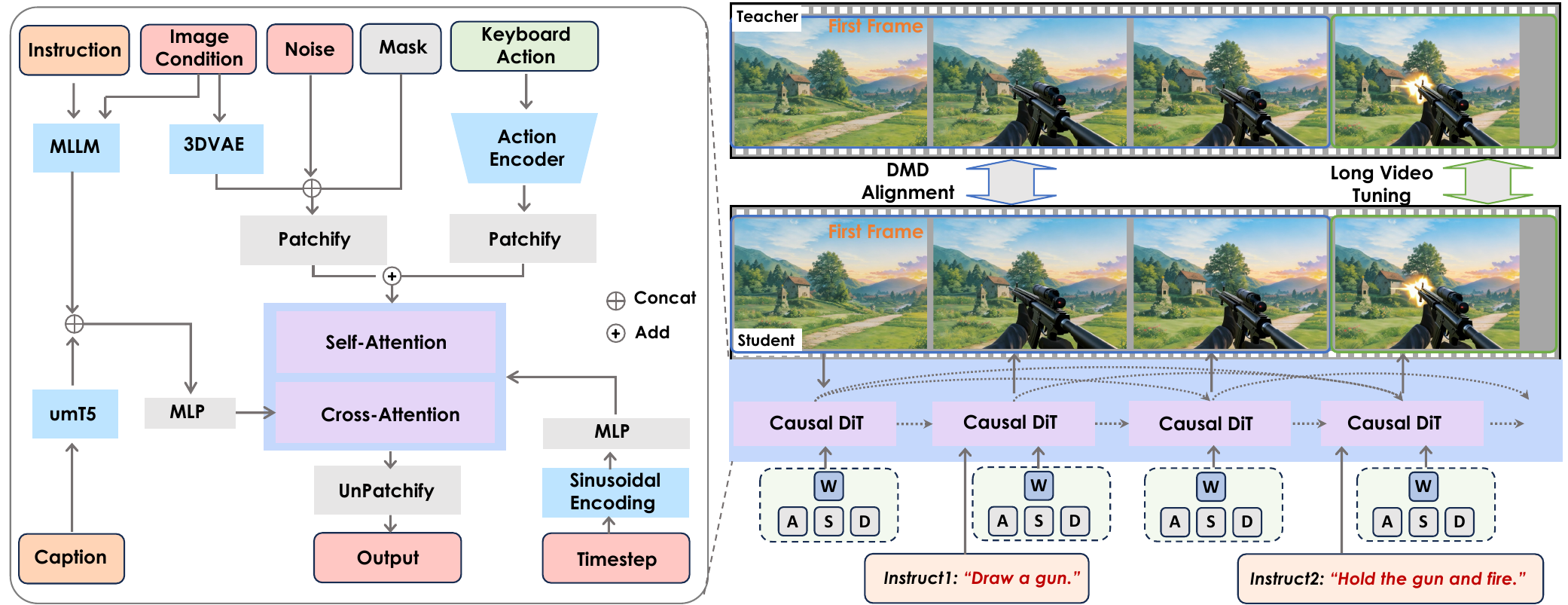}
    \caption{Model architecture of Hunyuan-GameCraft-2. Given a reference image and the corresponding action, the keyboard/mouse signal, and prompt-based instruction, we inject these options to the main architecture   (See Sec.~\ref{sec:arch}). During training and inference, we leverage self-forcing post-training for long-video extension(See Sec.~\ref{sec:train}), and KV-cache/recache for multi-action switching(See Sec.~\ref{sec:infer}). To maintain the long-term video quality, we design a randomized long video tuning scheme(See Sec.~\ref{sec:train}).}
    \label{fig:pipeline}
\end{figure*}

We present \textbf{Hunyuan-GameCraft-2}, an interactive game video model focusing on free-form instruction-based control. 
The overall framework is illustrated in Fig.~\ref{fig:pipeline}. In particular, our model unifies a natural action-injected causal architecture, image-conditioned autoregressive long video generation, and diverse multi-prompt interaction into a cohesive framework.
This section will introduce the model architecture, training, and inference procedures.

\subsection{Model Architecture}
\label{sec:arch}
The main architechture of our model is based on a 14B image-to-video mixture-of-experts (MoE) foundation video generation model~\cite{wan_wan_2025}. Our objective is to extend this image-to-video diffusion model into an action-controllable generator. As discussed in Sec.~\ref{sec:intro}, the action space includes both keyboard inputs and free-form text prompts.

For keyboard and mouse signal injection (W, A, S, D, ↑, ←, ↓, →, Space, etc.), we adopt the methodology from GameCraft-1~\cite{Hunyuan-GameCraft}, mapping these discrete action signals to continuous camera control parameters. During training, annotated camera parameters are encoded as Plücker embeddings~\cite{he2024cameractrl} and integrated into the model through token addition. At inference, user inputs are converted into camera trajectories to derive these parameters.

As for prompt-based interaction injection, 
we observe that the base model struggles to express certain interactive verbs, largely due to the higher semantic and spatial complexity of interaction texts compared to scene descriptions. Such texts are often tightly coupled with specific visual regions or object instances. To mitigate this, we leverage a multimodal large language model (MLLM)~\cite{wang2024qwen2} to extract, reason and inject interaction information to the main model, which can enrich interaction-related textual guidance, improving the model’s ability to differentiate between general text instructions and fine-grained interactive behaviors during training.
This camera-conditioned control, when combined with text-based scene and interaction inputs, forms a unified mechanism that enables Hunyuan-GameCraft-2 to navigate and interact seamlessly within its environment.

\subsection{Training Procedure}
\label{sec:train}
To achieve long-term and real-time interactive video generation, it's necessary to distill the foundational bidirectional model into a few-step causal generator.
In this work, we scale the comprehensive autoregressive distillation technique, Self-Forcing~\cite{huang2025selfforcingbridgingtraintest}, to a 14B Mixture-of-Experts (MoE) image-to-video model~\cite{wan_wan_2025}. This scheme is specifically tailored to enhance generation quality and efficiency for long video generation, which often features large and rapid scene variations. We introduce random extension tuning to mitigate error accumulation. The training process is organized into four stages: (1) Action-Injected Training, (2) Instruction-Oriented Supervised Fine-Tuning, (3) Autoregressive Generator Distillation, and (4) Randomized Long-Video Extension Tuning.

\begin{table*}[h]
\centering\small
\caption{\textbf{Detailed training configurations across different stages. CP denotes context parallelism.} }

\begin{tabular*}{\linewidth}{@{\extracolsep{\fill}} l r l r r}
\toprule
\textbf{Training Stage} & \textbf{Dataset} & \textbf{Data type} & \textbf{CP} & \textbf{\#iters}   \\ \midrule
Action-Injected Training     & 1M     & Game-play \& Render Video & 1   & 100k    \\
Instruction-Oriented SFT    & 150K        & Game-play \& Synthetic Video  & 1 & 20k    \\
Autoregressive Generator Distillation    & 200K         & Game-play \& Synthetic Video   & 4 & 10K \\ 
Randomized Long-Video Extension Tuning       & 100K & Game-play Long video & 4 & 3K  \\ \bottomrule
\end{tabular*}
\end{table*}

\subsubsection{Action-Injected Training}  
The primary objective of this stage is to establish a fundamental understanding of 3D scene dynamics, lighting, and physics. 
We load the pre-trained weights and finetune the model with the flow-matching
objective for architectural adaptation.
In order to improve the long-term consistency, we adopted a curriculum learning strategy. 
Specifically, we organized the training into three phases, exposing the model to video data of 45, 81, and 149 frames in 480p in sequence. 
This stepped approach allows the model to first solidify its understanding of short-term motion dynamics before gradually adapting its attention mechanisms to handle the complex dependencies required for longer-duration coherence. Besides, we randomly choose long and short captions during training, and concatenate interactive captions for interaction learning. This option will help the model to have an initial perception of the injection of interactive information. 

\subsubsection{Instruction-Oriented Supervised Fine-Tuning}
To enhance the model's interactive capabilities, we constructed a dataset of 150K samples by augmenting real-world footage with procedurally generated synthetic videos (details in Sec.~\ref{sec:data}). These synthetic sequences can provide high-fidelity supervision across diverse interaction types (e.g., state transitions, subject interactions), thereby establishing a tight correspondence between actions and their visual outcomes.
In the subsequent stage, we freeze the camera encoder's parameters and exclusively fine-tune the MoE experts. This process is designed to refine the model's alignment with semantic control cues.

\subsubsection{Autoregressive Generator Distillation}


For interactive world models, extending fixed-length video generators to high-quality autoregressive long-video generation is essential. Prior works have made preliminary attempts on long video generation~\cite{huang2025selfforcingbridgingtraintest,shin2025motionstream,voboril2025streamllm,yang2025longliverealtimeinteractivelong}.
Building upon the high- and low-noise MoE architecture and camera parameter injection, we introduce targeted adaptations to the attention mechanism and the distillation protocol. These modifications are specifically tailored to optimize performance within the autoregressive distillation process.

\noindent \textbf{Sink Token and Block Sparse Attention:} 
Previous arts~\cite{yin2025slowbidirectionalfastautoregressive,huang2025selfforcingbridgingtraintest} updates the KV cache for causal attention using a direct sliding-window approach. However, this can lead to a degradation in generation quality over time, as later steps cannot reference the initial conditioning frame, causing drift. Therefore, inspired by prior work~\cite{shin2025motionstream,voboril2025streamllm,yang2025longliverealtimeinteractivelong}, we designate the initial frame as a sink token, which is always retained in the KV cache. This modification serves two critical functions: firstly, it improves and stabilizes generation quality. Secondly, in our specific task, the sink token provides information about the coordinate system origin. This ensures that camera parameters injected during the autoregressive process remain consistently aligned with the initial frame, thereby avoiding the need for a recache at each autoregressive step due to shifts in the coordinate origin. Additionally, we employ Block Sparse Attention~\cite{guo2024blocksparse} for local attention, better suited for our autoregressive, block-wise generation process. Specifically, the target block being generated can attend to a set of preceding blocks. This local attention, combined with the aforementioned sink attention, constitutes the full KV cache, enhancing generation quality while also accelerating the generation speed.

\begin{figure}
    \centering
    \includegraphics[width=\linewidth]{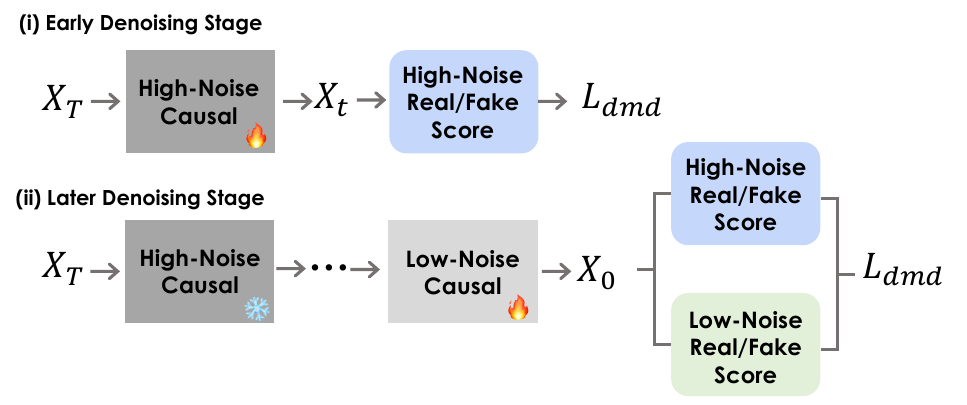}
    \caption{Distillation Schedule for Self-Forcing post training on the MoE Model.}
    \vspace{-6mm}
    \label{fig:dmd}
\end{figure}

\noindent \textbf{Distillation Schedule:} 
Due to the unique nature of the MoE architecture, the high-noise expert presents greater challenges in training and convergence than the low-noise expert~\cite{wan_wan_2025}, particularly during SFT or distillation. To address this, we assign distinct learning rates to each expert. Concurrently, we re-define the target list of denoising timesteps for distillation based on the noise level boundary that separates the two experts. This ensures that the teacher and student models maintain consistency in their selection of the high- or low-noise expert during the distillation process.


\begin{algorithm}[tp]
\caption{\textbf{Randomized Extended Long-Video Tuning}}
\label{alg:long_video_tuning}
\small
\begin{algorithmic}[1]
\Require Student \(G_{\theta}\), Real Score \(T_{\text{real}}\), Fake Score \(T_{\text{fake}}\), Dataset \(\mathcal{D}\), Cache size \(L\), Window \(K\), Max length \(N_{\text{max}}\), Timesteps \(\{t_1, \dots, t_T\}\)

\Loop

  \State \(V_{\text{gt}} \sim \textsf{Sample}(\mathcal{D})\) \Comment{Sample a ground truth video}
  \State \(N \sim \textsf{Sample}(\mathcal{U}(K, N_{\max}))\) \Comment{Randomize rollout length}
  \State \(V_{\text{pred}} \gets [V_{\text{gt}}[0]]\), \(\text{KV} \gets \emptyset\)   \Comment{Initialize with the first frame}
  \textcolor{MyTeal}{\algstep{Step 1: Autoregressive Rollout}}
  \For{\(j = 1\) to \(N/K\)}
    \State \(V_{\text{prev}} \gets \textsf{LastKFrames}(V_{\text{pred}}, K)\)
    \LineComment{Extend sequence autoregressively}
    \State \(V_{\text{chunk}}, \text{KV} \gets G_{\theta}(V_{\text{prev}}, \text{KV}, \text{sink\_token})\) 
    \State \textsf{Append}(\(V_{\text{pred}}\), \(V_{\text{chunk}}\))
  \EndFor
  \textcolor{MyTeal}{\algstep{Step 2: Randomized Window Sampling}}
  \State \(i \sim \textsf{Sample}(\mathcal{U}\{1, \dots, N-K+1\})\)
  \LineComment{Uniformly sample a predicted window}
  \State \(W \gets V_{\text{pred}}[i : i+K-1]\) 
  \textcolor{MyTeal}{\algstep{Step 3: Interleaved Forcing Logic}}
  \LineComment{Self-Forcing: Condition on predicted history}
  \State \(c_{\text{student}} \gets V_{\text{pred}}[i-1]\) 
  \LineComment{Teacher-Forcing: Condition on ground truth}
  \State \(c_{\text{teacher}} \gets V_{\text{gt}}[i-1]\) 
  \textcolor{MyTeal}{\algstep{Step 4: Distillation with Distinct Conditions}}
  \State \(t \sim \textsf{Sample}(\{t_1, \dots, t_T\})\), \(\epsilon \sim \mathcal{N}(0, \mathbf{I})\)
  \LineComment{Apply forward diffusion noise}
  \State \(x_t(W) \gets \textsf{NoiseSchedule}(W, t, \epsilon)\) 
  \LineComment{Calculate DMD loss}
  \State \(\mathcal{L} \gets \mathrm{DMD}\big( 
             T_{\text{fake}}(x_t(W), t, c_{\text{student}}), 
             T_{\text{real}}(x_t(W), t, c_{\text{teacher}}) 
             \big)\)
  \LineComment{Update generator parameters}
  \State \(\theta \gets \theta - \eta \nabla_{\theta}\mathcal{L}\) 
\EndLoop
\end{algorithmic}
\end{algorithm}

\subsubsection{Randomized Extended Long-Video Tuning}


Our approach to enabling long-form video generation is motivated by the observation that the foundation model, despite being pre-trained on short clips, implicitly captures the global visual data distribution.
Previous methods~\cite{cui2025self,yang2025longliverealtimeinteractivelong}, roll out long video sequences from a causal generator and apply distributional moment distance (DMD)~\cite{dmd,dmd2} alignment on the extended frames. This strategy effectively mitigates error accumulation during autoregressive generation.

Building upon this insight, we adopt a randomized extension tuning strategy using a dataset of long-form gameplay videos exceeding 10 seconds. 
In this stage, the model autoregressively rolls out $N$ frames, and contiguous $T$-frame windows are uniformly sampled to align the predicted and target distributions (either the ground truth or teacher priors). Furthermore, we randomly extend the predicted videos from the causal generator to varying lengths, promoting robustness across different temporal horizons.
In practice, while rolling out at window $W=V[i:i+K-1]$, the student generator uses sink token and KV cache and autogressively extend long video, and the fake score teacher model uses the last frame in the previous clean predicted chunk $V[i-1]$ as image condition; while the real score uses the ground truth frame in the original video. 

To mitigate the potential erosion of interactive capabilities inherent in few-step distillation, we adopt a training paradigm that interleaves self-forcing with teacher-forcing. The rationale for this approach is to compel the model to master state recovery and maintain temporal stability. Crucially, this is achieved by exposing it to diverse states at arbitrary points along the generation trajectory, rather than limiting such corrective training solely to the initial phase.




\subsection{Multi-turn Interactive Inference}
\label{sec:infer}

\begin{figure*}[h!]
    \centering
    \includegraphics[width=\linewidth]{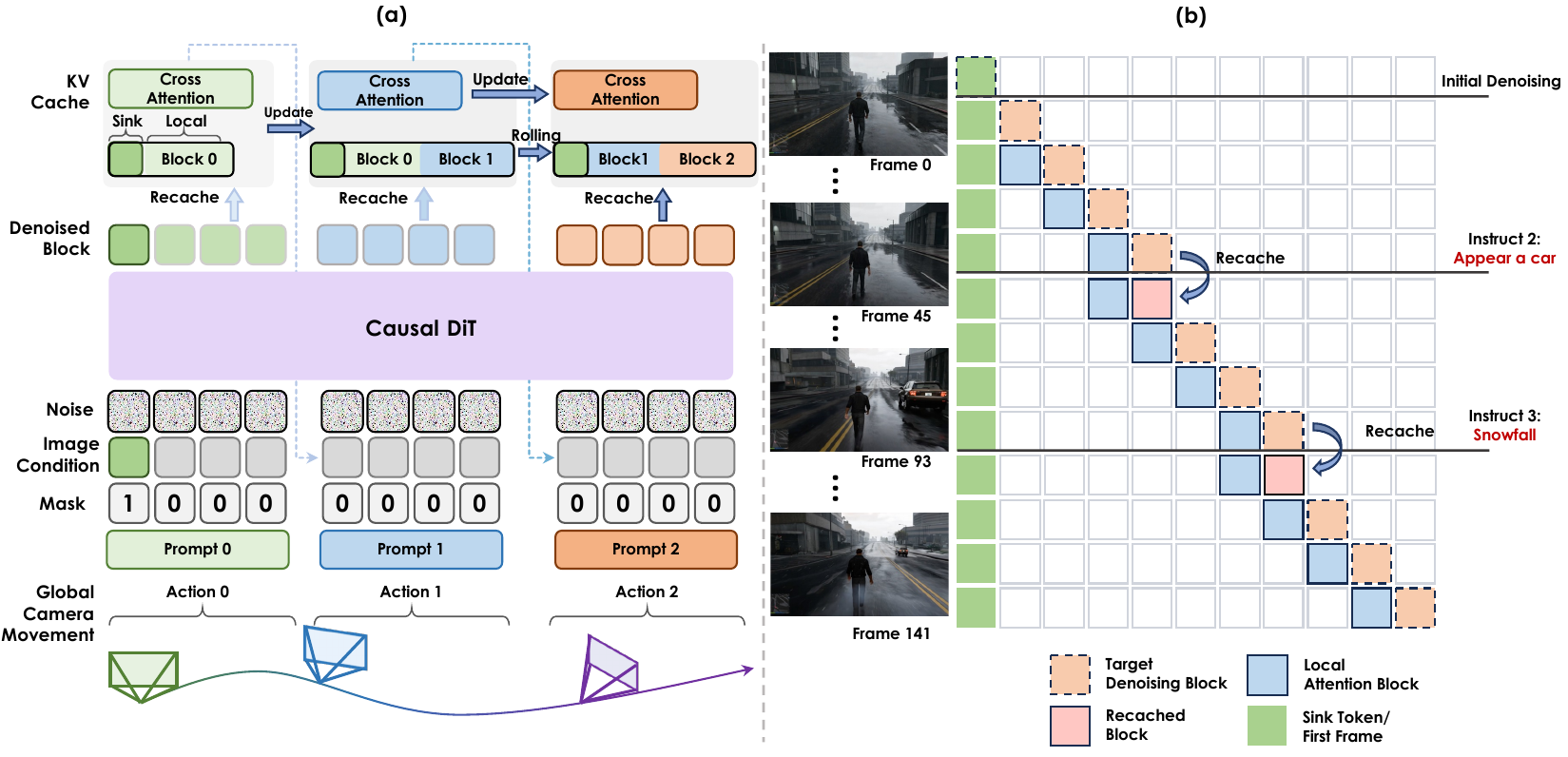}
    \caption{\textbf{Multi-turn Interactive Inference.} Figure 7.(a) illustrates our block-wise autoregressive inference pipeline for long video generation, along with the corresponding KV cache updating mechanism. The initial frame is retained as sink tokens at the start of the KV cache window, while local attention is derived from the recently generated blocks; The prompt recaching mechanism is depicted in Figure 7.(b), this strategy effectively enhances both the responsiveness and accuracy of the interaction when processing new prompts.}
    \label{fig:inference}
\end{figure*}

\noindent \textbf{Self-attention KV Cache}. To maintain consistency with the training strategy, our inference process employs a fixed-length self-attention KV cache with a rolling update mechanism to facilitate efficient autoregressive generation, as depicted in Fig.~\ref{fig:inference}. Specifically, sink tokens is permanently retained at the beginning of the cache window. The subsequent segment functions as a local attention window, maintaining the $N$ frames preceding the target denoising block throughout multi-turn interactions. The complete KV cache is composed of these sink tokens and the local attention component, which is implemented using block sparse attention. This design not only enhances autoregressive efficiency but also effectively prevents quality drift.

\noindent \textbf{ReCache Mechanism}. We employ a recache mechanism to enhance the accuracy and stability of multi-turn interactions in autoregressive long video generation. Upon receiving a new interaction prompt, the model extracts the corresponding interaction embeddings to recompute the last autoregressive block and update both the self-attention and cross-attention KV caches. This strategy provides precise historical context for the subsequent target block with minimal computational overhead, thereby ensuring accurate and responsive feedback to facilitate a smoother user experience.

\subsection{Real-time Interaction Acceleration}
To further accelerate inference and minimize latency, we incorporate several system-level optimizations:

\begin{itemize}
\item \textbf{FP8 Quantization} reduces memory bandwidth and leverages GPU acceleration while preserving visual quality;
\item \textbf{Parallelized VAE decoding} enables simultaneous latent-frame reconstruction, mitigating bottlenecks in long-sequence decoding;
\item \textbf{SageAttention}~\cite{zhang2025sageattentionaccurate8bitattention} replaces FlashAttention with an optimized quantized attention kernel for faster transformer computation; and
\item \textbf{Sequence parallelism }distributes video tokens across multiple GPUs, supporting efficient long-context generation.
\end{itemize}
Together, these techniques boost inference speed to 16 FPS, achieving real-time interactive video generation with stable quality and low latency.
\section{Experiments}
\label{sec:experiments}

\subsection{Model and Dataset Configurations.}
We compare our method against several SOTA image-to-video generation foundation models, including \textbf{HunyuanVideo}, \textbf{Wan2.2 A14B}, and \textbf{LongCatVideo}. For fairness, all baselines are evaluated under their recommended or commonly adopted inference configurations, detailed as follows:

\begin{itemize}
    \item \textbf{HunyuanVideo.}  
    We use the official configuration with the following settings: \texttt{FLOW\_SHIFT=7.0}, \texttt{EMBEDDED\_CFG\_SCALE=6.0}, 50 denoising steps, and with \texttt{flow\_reverse} and \texttt{i2v\_stability} enabled for enhanced temporal robustness.

    \item \textbf{Wan2.2 A14B.}  
We use the UniPC sampler, setting \texttt{sample\_shift=5.0}, \texttt{sample\_steps=40}, \texttt{boundary=0.900}, and using a dual-stage CFG with  scales of (3.5, 3.5) for both noise regimes.

    \item \textbf{LongCatVideo.}  
We use the default high-quality inference setup with a \texttt{guidance\_scale} of 4, 50 denoising steps, and enabled compilation optimizations for efficiency.

\end{itemize}

\paragraph{Resolution and Dataset.}  
To comprehensively evaluate controllable video generation, we constructed a test suite organized around three core interaction dimensions: \textbf{(1) Environmental Interactions}, \textbf{(2) Actor Actions}, and \textbf{(3) Entity and Object Appearances}. To support this framework, we curated a custom test set of \textbf{100 images}, covering a wide diversity of scenes (indoor/outdoor, natural/urban), lighting conditions, and visual styles (realistic, game-like, cartoon). Furthermore, we built specialized subsets for specific actions, such as an additional \textbf{20 images} of closed doors for evaluating the \textit{open door} action. For all evaluations, models are required to generate videos at a unified resolution of \textbf{832$\times$448} and a fixed length of \textbf{93 frames}.

\begin{table*}[t]
\centering\small
\caption{\textbf{Composition of our curated evaluation test set.} The data is hierarchically organized by high-level category, sub-category, and complexity level (Basic and Extended), with the number of prompts specified for each fine-grained subset.}

\label{tab:testset_refined_optimized}
\begin{tabular*}{\linewidth}{@{\extracolsep{\fill}} c l l l r}
\toprule
\textbf{Category} & \textbf{Sub-category} & \textbf{Level} & \textbf{Subset} & \textbf{Num} \\
\midrule

\multirow{4}{*}{\shortstack[c]{\textbf{Environmental}\\\textbf{Interactions}}}
    & \multirow{3}{*}{Weather} & \multirow{3}{*}{Basic}    
        & Snow       & \textbf{100} \\
    &                          &                           
        & Rain       & \textbf{100} \\
    &                          &                           
        & Lightning  & \textbf{100} \\
    \cmidrule(lr){2-5}
    & Physical event           & Extended                  
        & Explosion  & \textbf{100} \\
\midrule

\multirow{6}{*}{\shortstack[c]{\textbf{Actor}\\\textbf{Actions}}}
    & \multirow{3}{*}{Primitive actions} & \multirow{3}{*}{Basic} 
        & Draw gun               & \textbf{100} \\
    &                                    &                        
        & Draw knife             & \textbf{100} \\
    &                                    &                        
        & Take out torch         & \textbf{100} \\
    \cmidrule(lr){2-5}
    & \multirow{3}{*}{Composite actions} & \multirow{3}{*}{Extended} 
        & Draw and fire gun      & \textbf{100} \\
    &                                    &                           
        & Take out and operate phone & \textbf{100} \\
    &                                    &                           
        & Open door              & \textbf{20}  \\
\midrule

\multirow{10}{*}{\shortstack[c]{\textbf{Entity \& Object}\\\textbf{Appearances}}}
    & \multirow{5}{*}{Animals}    & Basic    
        & Cat      & \textbf{25} \\
    &                             & Basic    
        & Dog      & \textbf{25} \\
    &                             & Basic    
        & Wolf     & \textbf{25} \\
    &                             & Basic    
        & Deer     & \textbf{25} \\
    &                             & Extended 
        & Dragon   & \textbf{100} \\
    \cmidrule(lr){2-5}
    & \multirow{4}{*}{Vehicles}   & \multirow{4}{*}{Basic}    
        & Red SUV              & \textbf{25} \\
    &                             &                           
        & Blue truck           & \textbf{25} \\
    &                             &                           
        & Yellow sports car    & \textbf{25} \\
    &                             &                           
        & Black off-road car   & \textbf{25} \\
    \cmidrule(lr){2-5}
    & Humans                      & Extended 
        & Human appearances & \textbf{100} \\
\bottomrule
\end{tabular*}
\end{table*}

\subsection{Evaluation Metrics}

To comprehensively evaluate the performance of our model in video generation, we employ two complementary families of metrics: general video-quality metrics and our interaction-focused evaluation suite, \textbf{InterBench}. The general metrics assess foundational aspects such as visual fidelity, temporal consistency, and motion realism, providing a baseline measurement of overall video quality. However, such metrics alone are insufficient for capturing the causal structure, action execution, and state transitions that are essential to interactive video generation. To bridge this gap, InterBench introduces \textbf{six interaction-centric dimensions}, each specifically designed to assess core properties of interactive behavior—including interaction completeness, action effectiveness, causal coherence, and physical plausibility. Together, these two metric families form a holistic and rigorous evaluation framework for interactive video models.

\begin{table*}[!ht]
\centering\small
\setlength{\tabcolsep}{1.4pt}
\caption{\textbf{Quantitative comparison of recent controllable video generation models.} 
GameCraft-2 achieves a strong overall balance across visual quality, temporal consistency, camera-control accuracy, and efficiency.}
\label{tab:model_eval}

\resizebox{\linewidth}{!}{
\begin{tabular}{lcccc cc c c}
\toprule
\multirow{2}{*}{\textbf{Model}} & 
\multicolumn{4}{c}{\textbf{Visual Quality}} & 
\multicolumn{1}{c}{\textbf{Temporal}} &
\multicolumn{2}{c}{\textbf{RPE}} &
\multirow{2}{*}{\thead{\textbf{FPS}$\uparrow$ }}
\\
\cmidrule(lr){2-5} \cmidrule(lr){6-6} \cmidrule(lr){7-8}
& {\textbf{FVD}$\downarrow$} & {\textbf{Image Quality}$\uparrow$} & {\textbf{Dynamic Average}$\uparrow$} & {\textbf{Aesthetic}$\uparrow$} & {\textbf{Temporal Consistency}$\uparrow$}  & {\textbf{Trans}$\downarrow$} & {\textbf{Rot}$\downarrow$} \\
\midrule
\textbf{GameCraft} & \textbf{1554.2} & 0.69 & \textbf{67.2} & 0.67 & 0.95  & \textbf{0.08} & 0.20 & 0.25\\
\textbf{GameCraft-PCM} & 1883.3 & 0.67 & 43.8 & 0.65 & 0.93  & \textbf{{0.08}} & {0.20}  &  {6.6}\\
\textbf{Matrix-Game} & 2260.7 & {0.72} & 31.7 & 0.65 & 0.94  & 0.18 & 0.35 & 0.06 \\
\textbf{Matrix-Game-2.0} & 1920.6 & 0.62 & 20.5 & 0.49 & 0.84 & \textbf{0.08} & 0.25 & \textbf{16} \\
\midrule
\rowcolor{gray!15}
\textbf{GameCraft-2}& 1856.3 & 0.70 & 45.2 & \textbf{0.71} & \textbf{0.96} & \textbf{0.08} & \textbf{0.17} & \textbf{16} \\ 
\bottomrule
\end{tabular}
}
\end{table*}

\subsubsection{General Metrics} To provide a comprehensive assessment of our model, we adopt a diverse set of evaluation metrics. For \textbf{video realism}, we use the Fréchet Video Distance (FVD)~\cite{unterthiner2019fvd}, which jointly captures spatial fidelity and temporal dynamics. \textbf{Visual quality} is quantified using Image Quality and Aesthetic scores, reflecting both low-level perceptual clarity and higher-level visual appeal. We further measure temporal consistency to evaluate cross-frame coherence and detect artifacts such as flickering or structural instability. For \textbf{dynamic performance}, we adapt the Dynamic Degree metric from VBench~\cite{huang2024vbench}. Instead of the original binary motion classification, we directly report absolute optical flow magnitudes, referred to as Dynamic Average. This continuous formulation provides a more nuanced characterization of motion intensity and naturalness.

For \textbf{interactive camera control performance}, we employ a multi-faceted evaluation protocol. We use the \textbf{Relative Pose Error }(\textbf{RPE trans} and\textbf{ RPE rot}) to measure trajectory control accuracy, computed after applying a Sim3 Umeyama alignment between the predicted reconstructed trajectory and the ground truth. This alignment removes scale and global pose discrepancies, allowing RPE to specifically reflect local motion fidelity and frame-to-frame control precision. By examining both translational and rotational components, the metric provides a clearer view of how accurately the model responds to interactive inputs and how reliably it maintains the intended motion trajectory.

\subsubsection{InterBench: Benchmarking Action-Level Interaction in Video Generation}
\label{sec:interbench}

To rigorously assess \textit{action-level interaction} in generated videos, we propose \textbf{InterBench}, a six-dimensional evaluation protocol tailored to interactive video generation. Instantiated using a vision-language model (VLM) as an automatic evaluator, InterBench is designed to measure not only whether an interaction is triggered, but also its fidelity, smoothness, and physical plausibility over time. The six core dimensions are defined below. For a comprehensive discussion of the protocol,  please refer to Appendix~\ref{sec:appendix_interbench}.

\begin{figure*}[h]
    \centering\includegraphics[width=\linewidth]{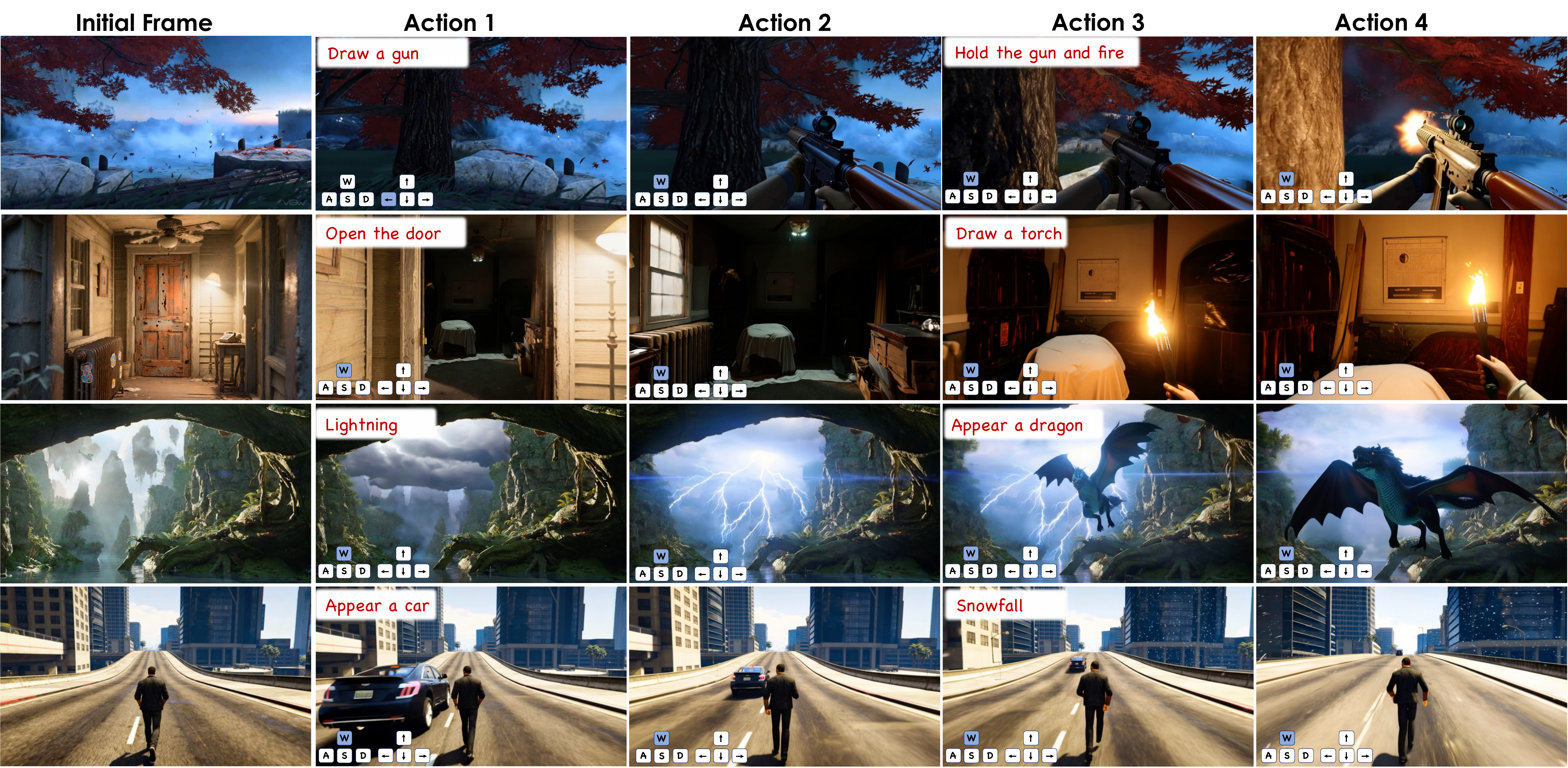}
    \caption{\textbf{Inference results by Hunyuan-GameCraft-2 on multi-action control.}}
    \label{fig:multi}
\end{figure*}
\begin{figure*}[h]
    \centering\includegraphics[width=\linewidth]{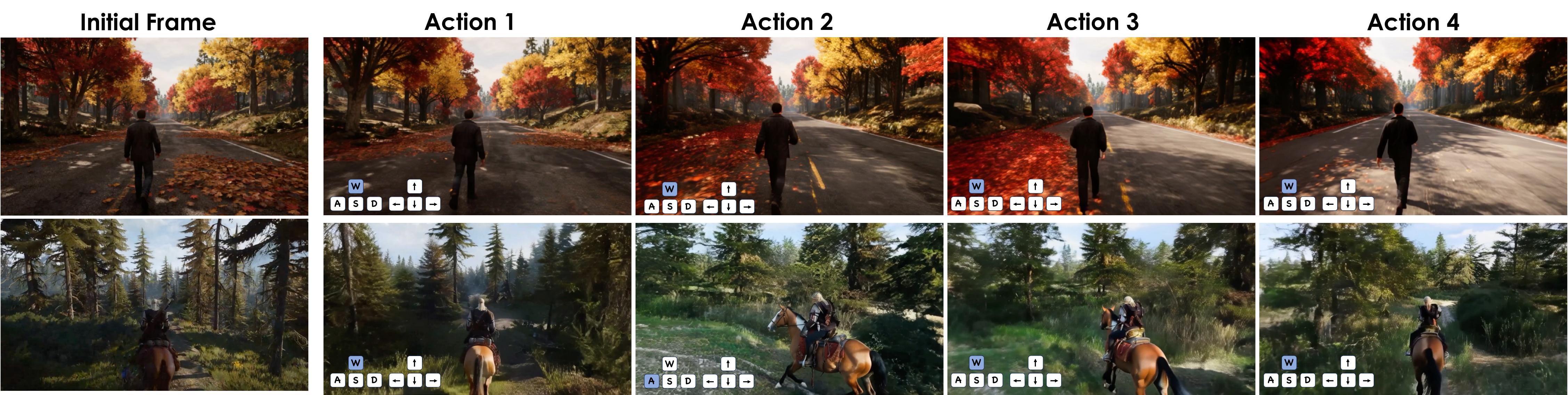}
    \caption{\textbf{Inference results by Hunyuan-GameCraft-2 on the third-perspective long-term game video
generation.}}
    \label{fig:thirdvidew}
\end{figure*}

\begin{enumerate}[leftmargin=18pt]

    \item \textbf{Interaction Trigger Rate.}
    A fundamental binary metric that assesses whether the requested interaction was successfully initiated. This serves as a gateway check, separating cases where the model completely ignored the prompt from those where it attempted the action.

    \item \textbf{Prompt--Video Alignment.}
    Evaluates the semantic fidelity between the video and the full prompt. This dimension has two facets: \textit{static alignment} (maintaining the scene's context and objects) and \textit{dynamic alignment} (executing the correct action as specified).

    \item \textbf{Interaction Fluency.}
    Measures the temporal naturalness and visual coherence of the interaction process. It penalizes temporal artifacts such as sudden jumps, flickering, or object teleportation that break the illusion of continuous motion and a stable timeline.

    \item \textbf{Interaction Scope Accuracy.}
    Examines whether the spatial extent of an interaction's effects is appropriate. It ensures that global events (like \textit{weather changes}) affect the entire scene, while local actions (like \textit{``lighting a torch''}) have a contained but realistic area of influence.

    \item \textbf{End-State Consistency.}
    Assesses whether the interaction converges to a stable and correct final state that persists until the end of the video. This distinguishes successful actions from those that are only partially completed or whose effects vanish prematurely.

    \item \textbf{Object Physics Correctness.}
    Evaluates the physical plausibility of interacting agents and objects. This includes maintaining the structural integrity of rigid bodies (no unnatural deformation), ensuring realistic motion kinematics, and preserving correct contact relationships (e.g., no penetration between hands and objects).
    
\end{enumerate}

\begin{figure*}
    \centering
    \includegraphics[width=\linewidth]{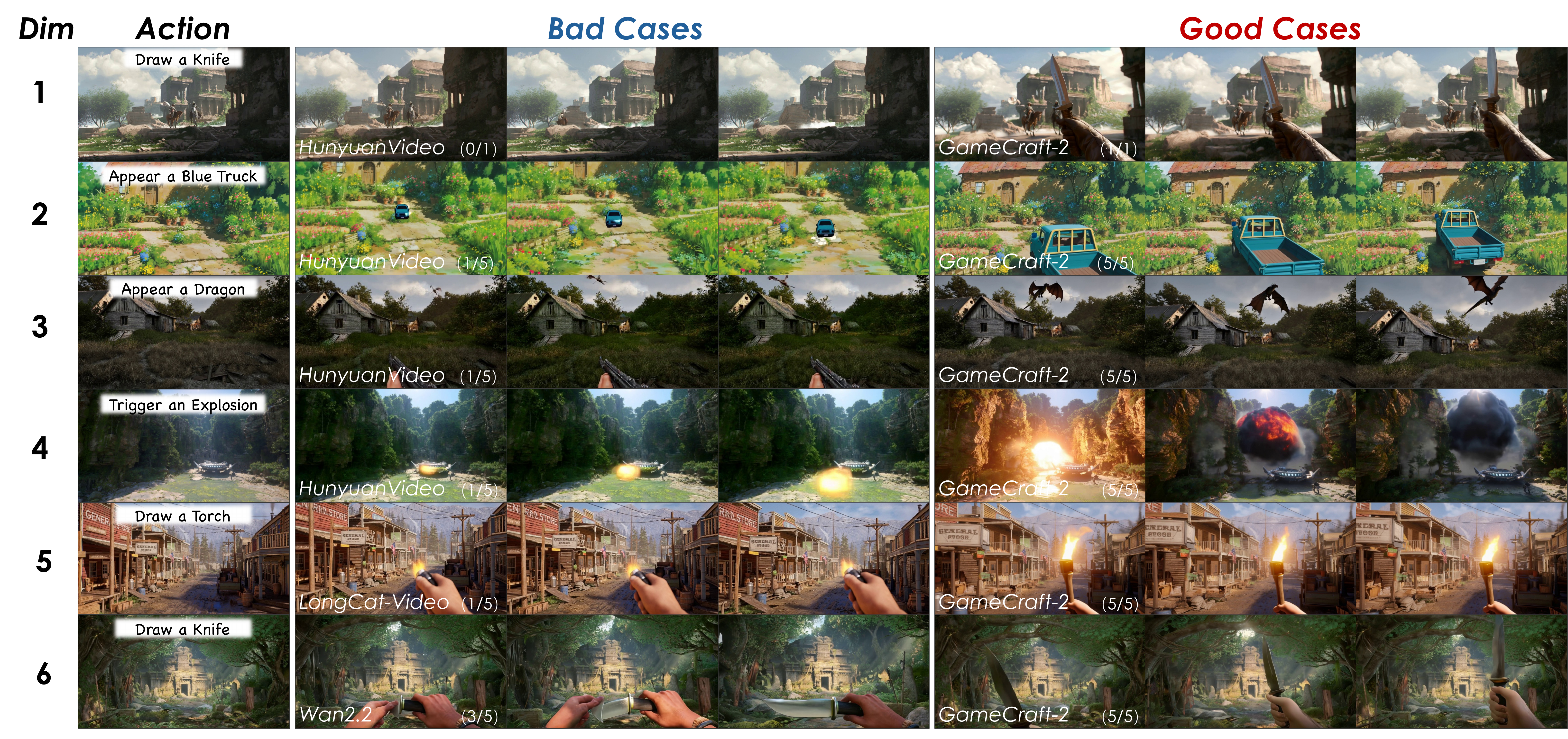}
    \caption{\textbf{Qualitative examples illustrating the six dimensions of our InterBench protocol.} Each row showcases a distinct evaluation dimension: \textbf{(1) Interaction Trigger Rate, (2) Prompt--Video Alignment, (3) Interaction Fluency, (4) Interaction Scope Accuracy, (5) End-State Consistency}, and \textbf{(6) Object Physics Correctness}. For each dimension, we present a high-scoring example from our model (right) against a low-scoring failure case from a baseline (left), demonstrating the rating scale used in our evaluation. GameCraft-2 consistently achieves higher ratings across all dimensions.}
    \label{fig:good_bad_case}
\end{figure*}

\paragraph{Scoring Protocol.}
Each video is evaluated against the InterBench dimensions using a discrete, ordinal scoring system. Specifically, \textbf{Interaction Trigger Rate} is assessed with a binary value (success/failure), while the remaining five dimensions receive multi-level ordinal scores to capture varying degrees of interaction quality. These per-video scores are then averaged to produce a score for each interaction category. A final, global InterBench score is obtained by aggregating these category-level results. This hierarchical scoring protocol enables both fine-grained analysis of specific failure modes and high-level comparison of interactive capabilities across different models.

\paragraph{Prompt Design.}  
To ensure fair and controlled evaluations, we designed a standardized, two-part prompt strategy. This approach constructs two complementary components for each test image: an \textit{\textbf{interaction prompt}} to specify the dynamic target action or event, and a \textit{\textbf{base prompt}} to describe static scene attributes  and anchor the generation process to the input image's appearance. During inference, these two prompts are concatenated into a single conditioning sentence and fed directly to each model. This decoupled design not only ensures that all models receive identical instructions for fair comparison but, critically, it also enables controlled evaluations by allowing us to systematically vary interaction instructions while keeping the visual context constant.

\begin{figure*}[h!]
    \centering
    \includegraphics[width=\linewidth]{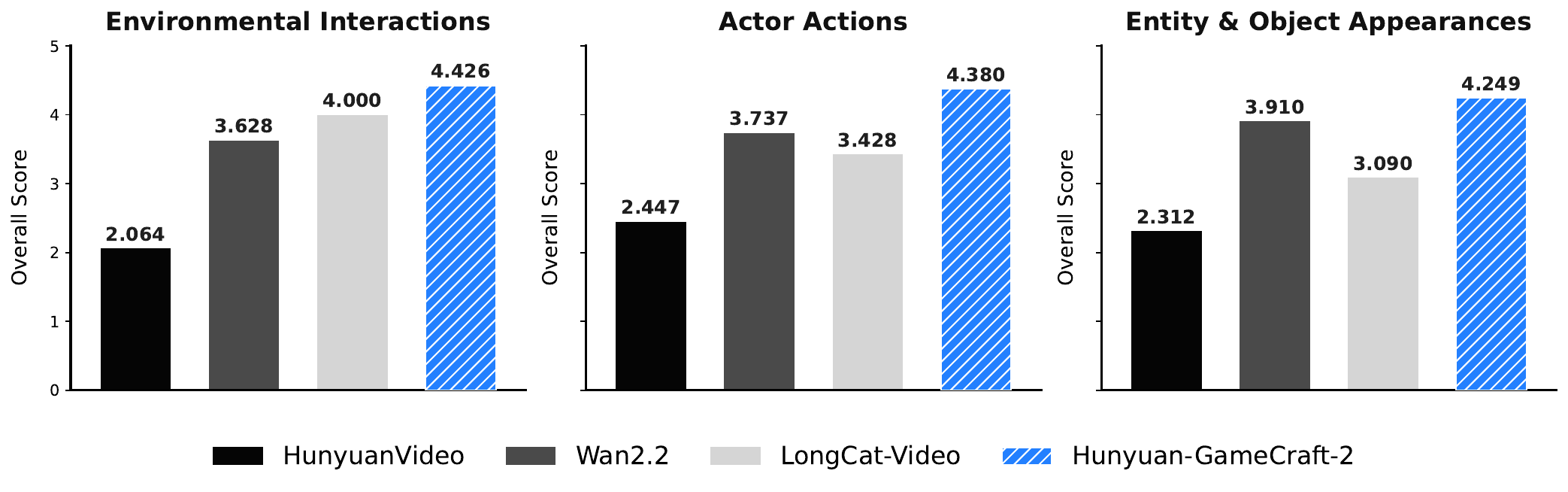}
    \vspace{-7mm}
\caption{\textbf{Comparison of Environmental Interactions with Baseline Models.} 
Qualitative results showing the fidelity and consistency of environment-level effects. Our approach better preserves global influence and temporal stability.}

    \label{fig:overall_comparison}
\end{figure*}

\begin{figure*}[h!]
    \centering
    \includegraphics[width=\linewidth]{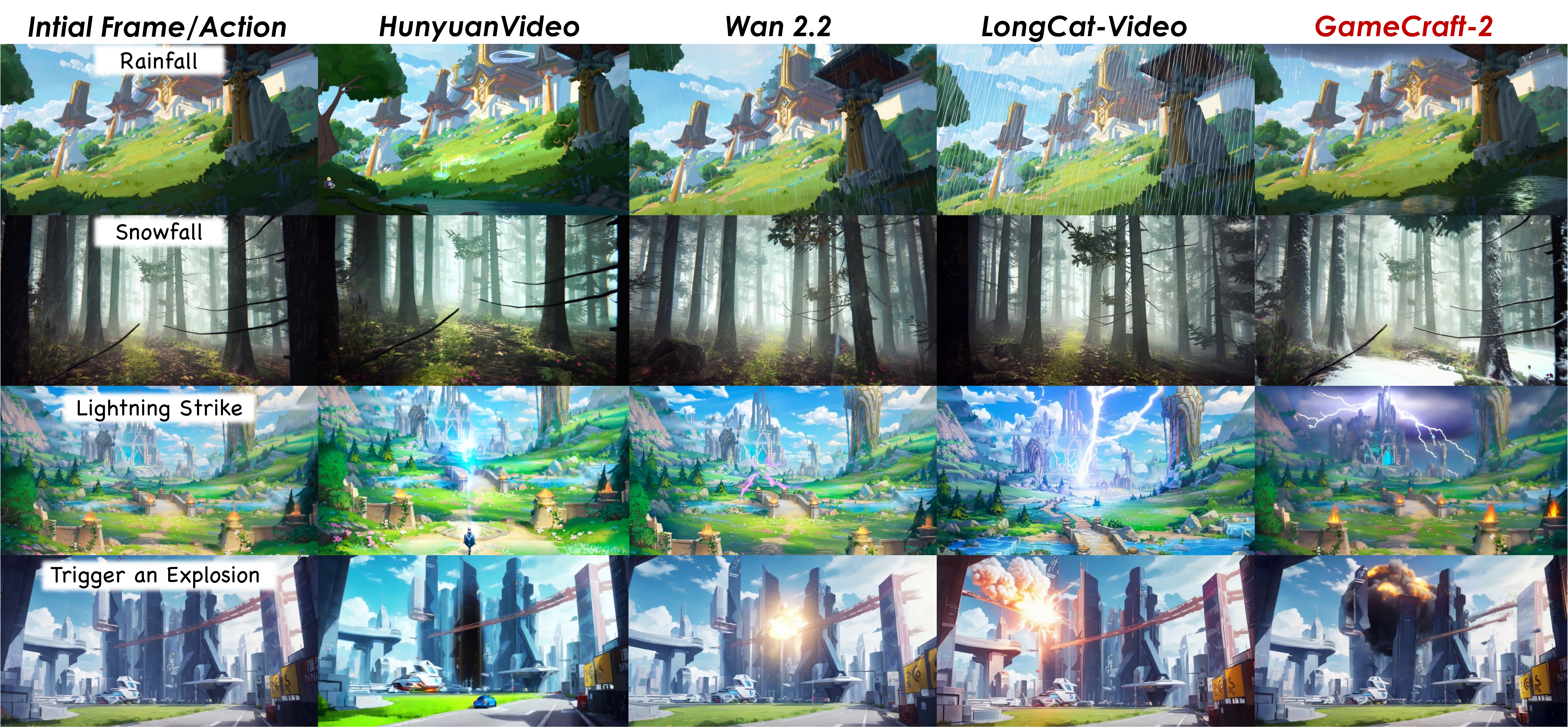}
\caption{\textbf{Comparison of Environmental Interactions with Baseline Models.} 
Qualitative results showing the fidelity and consistency of environment-level effects. Our approach better preserves global influence and temporal stability.}

    \label{fig:env_scores}
\end{figure*}

\begin{figure*}[h!]
    \centering
    \includegraphics[width=\linewidth]{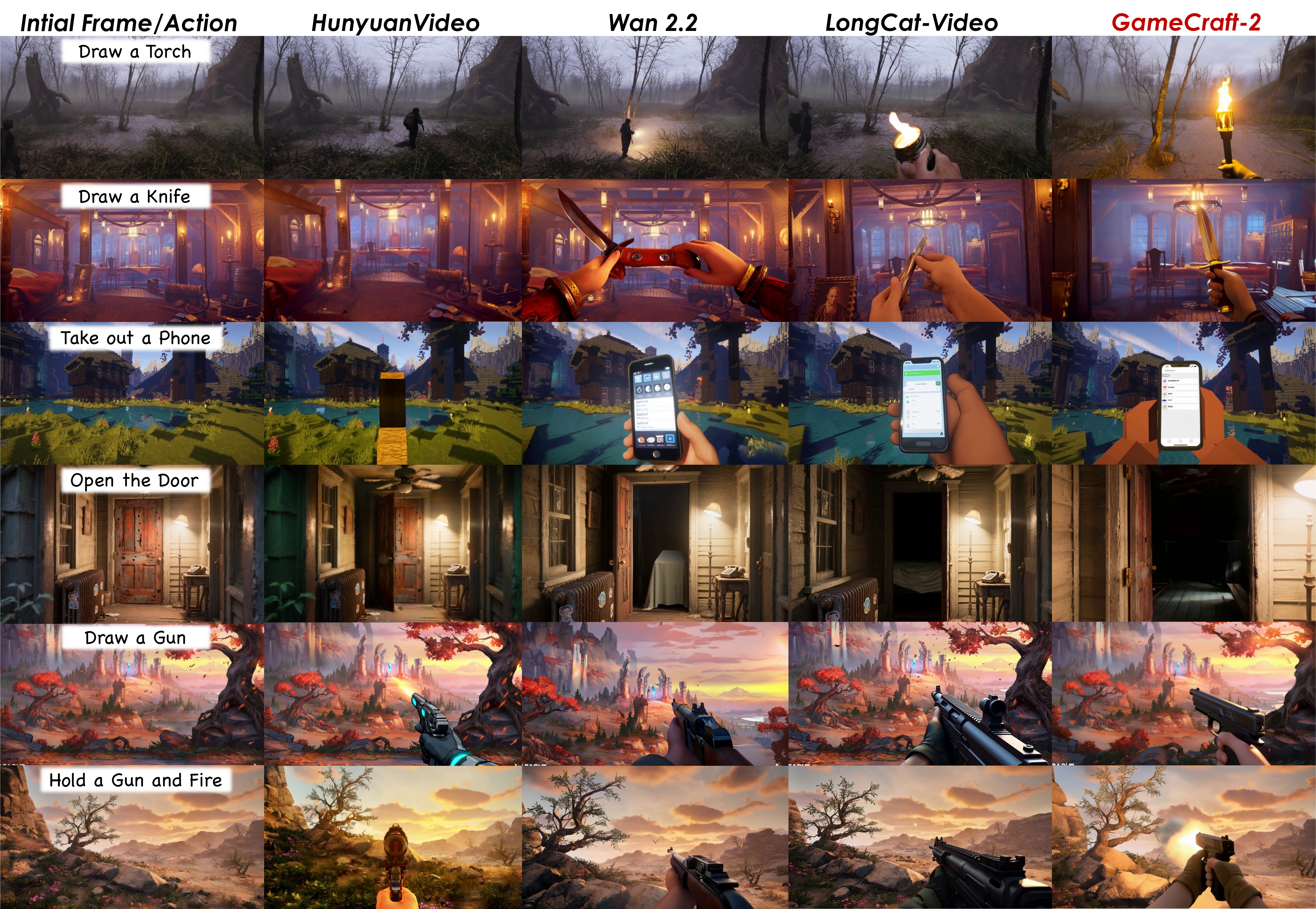}
    \vspace{-6mm}
    \caption{\textbf{Comparison of Actor-Action Interactions with Baseline Models.} 
Visual comparisons illustrating the quality of action-level interactions across representative prompts. Our method produces more coherent and physically consistent actions than all baselines.}

    \label{fig:actor_scores}
\end{figure*}

\begin{figure*}[h!]
    \centering
    \includegraphics[width=\linewidth]{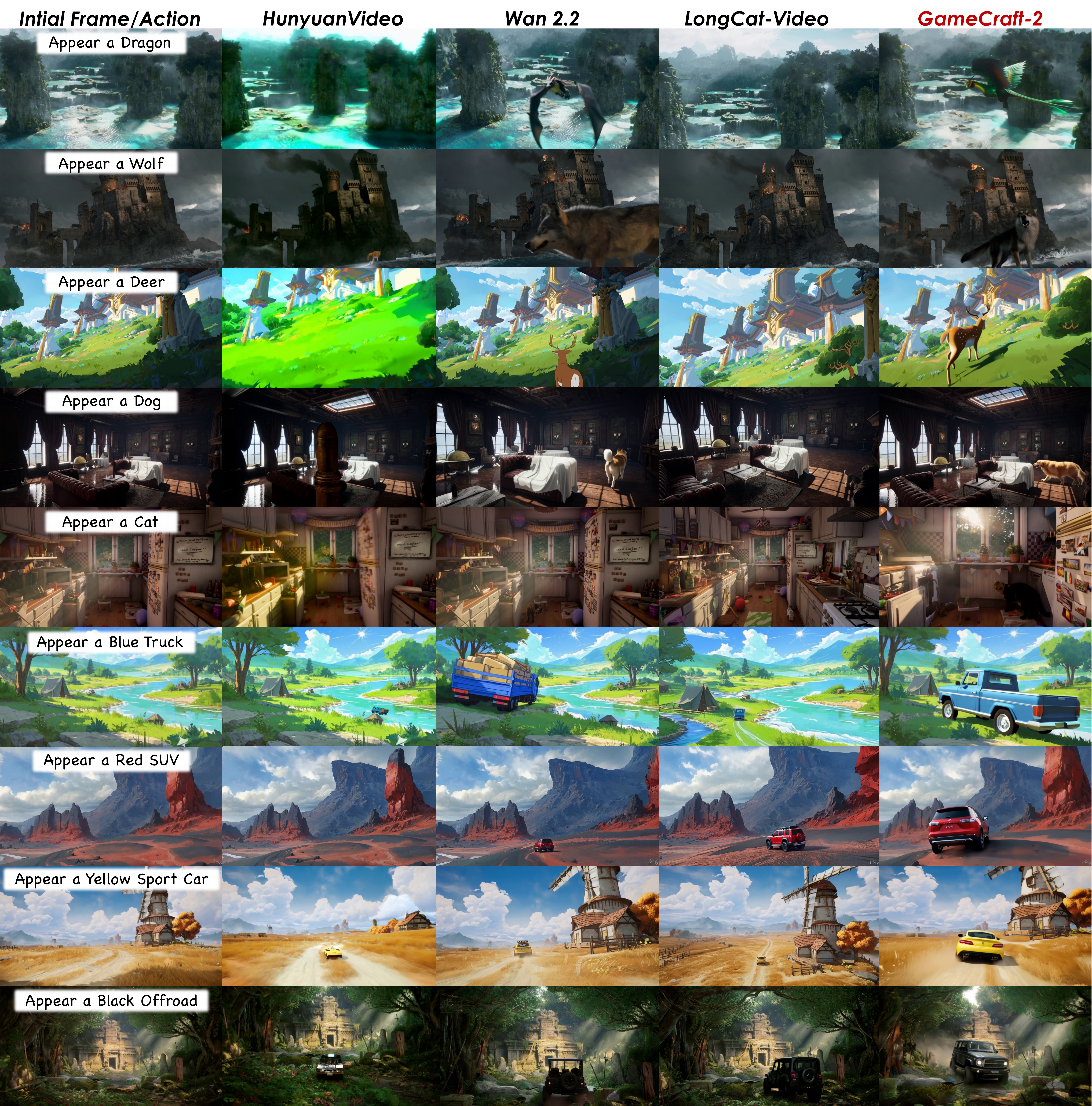}
        \vspace{-6mm}
\caption{\textbf{Comparison of Entity and Object Appearance Interactions with Baseline Models.} 
Visual comparisons of object emergence and interaction correctness. Our method delivers more accurate, stable, and physically plausible object behaviors.}

    \label{fig:entity_scores}
\end{figure*}

\subsection{Interaction Evaluation}

\paragraph{Quantitative Results on Interaction Evaluation}

We present the quantitative results for the three interaction categories in Table~\ref{tab:effect}. The evaluation follows our proposed InterBench protocol (Sec.~\ref{sec:interbench}), structured around its six core dimensions: \textbf{Trigger}, \textbf{Align}, \textbf{Fluency}, \textbf{Scope}, \textbf{EndState}, and \textbf{Physics}. To provide a single aggregated metric for comparison, we also compute a weighted \textit{Overall} score:

\[
\begin{split}
    \text{Overall} = \Big( & 5 \times \text{Trigger} + \text{Align} + \text{Fluency} \\
    & + \text{Scope} + \text{EndState} + \text{Physics} \Big) / 6.
\end{split}
\]

\begin{table*}[!ht]
\centering\small
\caption{\textbf{Quantitative performance evaluation}, against state-of-the-art competitors on the InterBench protocol. Scores are presented across six key interaction dimensions, with our model's superior results highlighted.}
\label{tab:effect}
\begin{tabular*}{\linewidth}{@{\extracolsep{\fill}} l l cccccc}
\toprule
\textbf{Category} & \textbf{Method} & \textbf{Trigger} & \textbf{Align} & \textbf{Fluency} & \textbf{Scope} & \textbf{EndState} & \textbf{Physics}  \\
\midrule
\multirow{4}{*}{\parbox{2.5cm}{\centering\textbf{Environmental \\ Interactions}}}
    & \textbf{Wan2.2 A14B}     & 0.799 & 3.511 & 3.579 & 3.722 & 3.951 & 3.008  \\
    & \textbf{LongCat-Video}   & 0.897 & 3.963 & 3.777 & 4.188 & 4.377 & 3.210  \\
    & \textbf{HunyuanVideo}    & 0.490 & 1.950 & 1.940 & 2.065 & 2.308 & 1.670  \\

    & \textbf{GameCraft-2} & \textbf{0.962} & \textbf{4.342} & \textbf{4.247} & \textbf{4.578} & \textbf{4.688} & \textbf{3.893}  \\
    
\midrule
\multirow{4}{*}{\parbox{2.5cm}{\centering\textbf{Actor \\ Actions}}} 
    & \textbf{Wan2.2 A14B}     & 0.836 & 3.490 & 3.488 & 4.036 & 4.054 & 3.175  \\
    & \textbf{LongCat-Video}   & 0.806 & 3.089 & 3.005 & 3.832 & 3.771 & 2.839  \\
    & \textbf{HunyuanVideo}    & 0.587 & 2.147 & 2.202 & 2.717 & 2.748 & 1.931  \\

    & \textbf{GameCraft-2} & \textbf{0.983} & \textbf{4.087} & \textbf{4.191} & \textbf{4.576} & \textbf{4.686} & \textbf{3.828} \\
\midrule
\multirow{4}{*}{\parbox{2.5cm}{\centering\textbf{Entity \& Object \\ Appearances}}} 
    & \textbf{Wan2.2 A14B}     & 0.874 & 3.943 & 3.545 & 4.281 & 4.265 & 3.054  \\
    & \textbf{LongCat-Video}   & 0.712 & 3.050 & 2.758 & 3.340 & 3.482 & 2.352  \\
    & \textbf{HunyuanVideo}    & 0.607 & 2.037 & 1.870 & 2.736 & 2.734 & 1.462  \\

    & \textbf{GameCraft-2} & \textbf{0.944} & \textbf{4.292} & \textbf{3.978} & \textbf{4.410} & \textbf{4.514} & \textbf{3.578} \\
\bottomrule
\end{tabular*}
\end{table*}

A category-wise quantitative analysis (Table~\ref{tab:effect}) reveals that GameCraft-2's superiority begins with its exceptionally high success rate in initiating interactions. The model achieves \texttt{Trigger} scores of 0.962 for \textbf{Environmental Interactions} and a near-perfect 0.983 for \textbf{Actor Actions}, far surpassing all baselines. 
Beyond successful initiation, GameCraft-2 excels in modeling the fidelity of these interactions. This is particularly evident in its physical realism, where it outperforms the next-best model by margins of 0.683 in \texttt{Physics} for Environmental Interactions and over 0.52 in \textbf{Entity \& Object Appearances}. 
Furthermore, it demonstrates substantial gains in temporal coherence and final state stability, with \texttt{Fluency} and \texttt{EndState} scores improving by +0.70 and +0.63, respectively, for Actor Actions. 
Collectively, these results underscore GameCraft-2's advanced capability not only to trigger interactions reliably but also to render them with high fidelity in semantics, dynamics, and physical consistency.

\paragraph{Qualitative Analysis}
For an intuitive demonstration of performance differences, we present a qualitative comparison in Figs.~\ref{fig:qual_env} -\ref{fig:qual_entity}. The results clearly highlight the superior performance of Hunyuan-GameCraft-2 over baseline models. Baselines frequently exhibit noticeable deficiencies when handling complex interactions. For instance, environmental effects  often lack dynamic evolution and realistic lighting interactions. Actor actions  are commonly plagued by object deformation, motion incoherence, and inaccurate hand-object contact. Furthermore, newly generated entities  tend to suffer from identity drift, unstable geometry, and poor integration with the scene.
In contrast, Hunyuan-GameCraft-2 demonstrates substantially higher fidelity and consistency across all interaction categories. In {Environmental Interactions}, its generated effects, such as snowfall, achieve global coverage and dynamic accumulation, rendering them more physically plausible. For {Actor Actions}, Hunyuan-GameCraft-2 produces more coherent action sequences, enabling characters to stably grasp and precisely manipulate objects while ensuring stable final states. In {Entity \& Object Appearances}, the model consistently maintains the structural integrity and identity of objects, seamlessly integrating them into the scene's lighting and perspective. {Crucially, this robustness extends to concepts outside our specific training categories; for instance, the model adeptly handles interactions involving a ``Phone'' or the appearance of a ``Dragon'', showcasing strong generalization capabilities.} Collectively, these qualitative examples not only corroborate our quantitative findings but also concretely showcase Hunyuan-GameCraft-2's robust capability to generate semantically accurate, temporally coherent, and physically plausible videos of complex interactions.

\paragraph{Generalization Beyond Training Distribution}
A notable strength of Hunyuan-GameCraft-2 lies in its ability to generalize interactive dynamics beyond the specific entities and scenarios present in the training data. Rather than merely memorizing visual patterns, the model internalizes the underlying structure of interaction---how agents initiate, propagate, and complete state transitions. As a result, Hunyuan-GameCraft-2 can robustly handle previously unseen subjects and objects. For example, although our dataset contains no instances of a ``man'' suddenly appearing, a ``dragon'' emerging, or an actor ``taking out a phone,'' the model successfully produces coherent and physically plausible interactions for all of these cases. It achieves this by leveraging learned principles of object emergence, action-driven causality, and hand–object coordination, enabling it to map novel concepts onto familiar interaction patterns. This demonstrates that the model has acquired a transferable representation of interactive processes, allowing it to extrapolate to open-domain scenarios far beyond the scope of its training distribution.
\begin{figure}[h!]
    \centering
    \includegraphics[width=\linewidth]{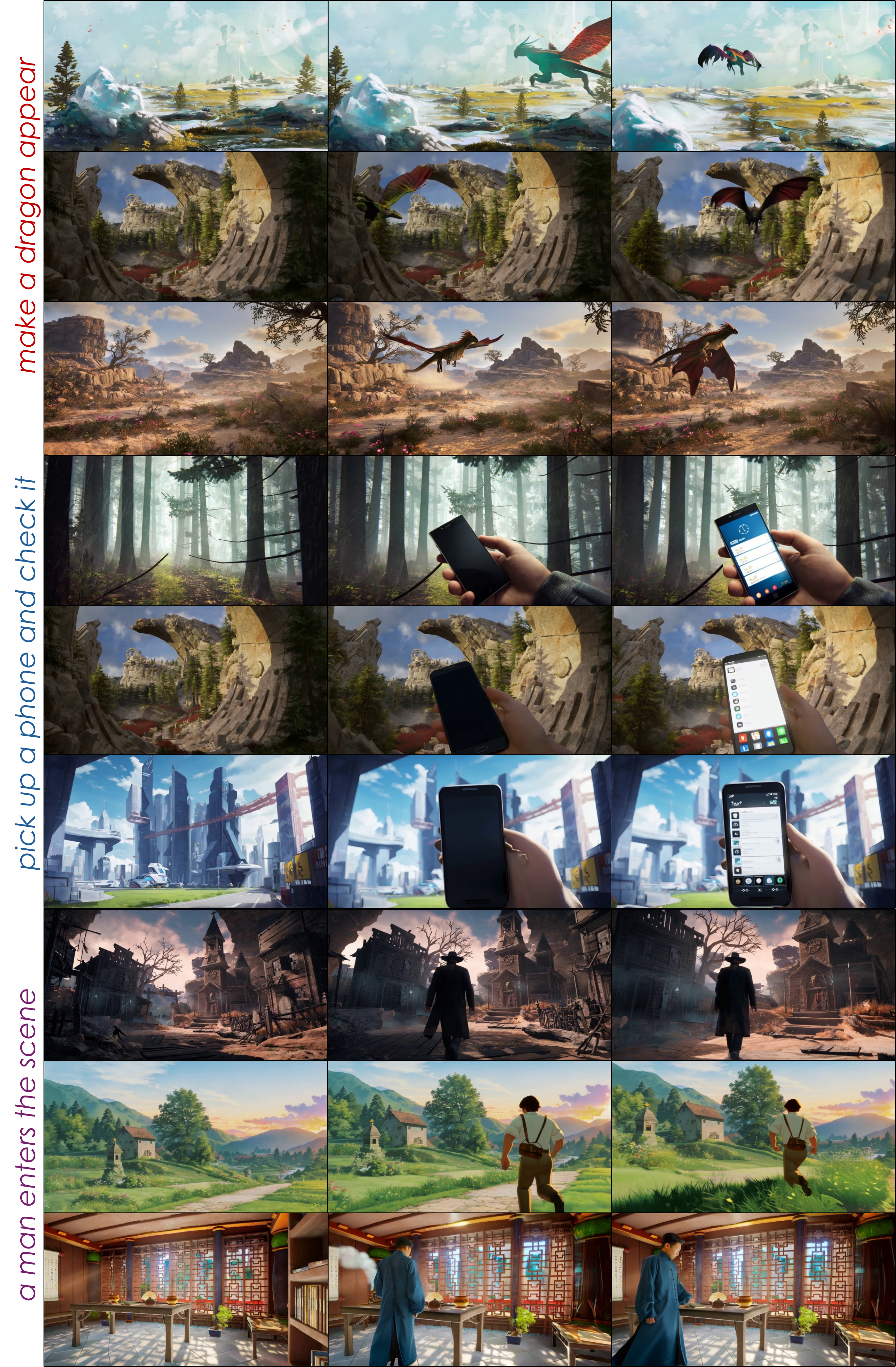}
    \caption{\textbf{Generalization to Unseen Entities and Actions.}
    Examples showing that Hunyuan-GameCraft-2 successfully handles interactions involving previously unseen subjects and objects The model produces coherent and physically plausible state transitions despite these cases being absent from the training data.}
    \label{fig:generalization_examples}
\end{figure}

\begin{figure}[h!]
    \centering
    \includegraphics[width=\linewidth]{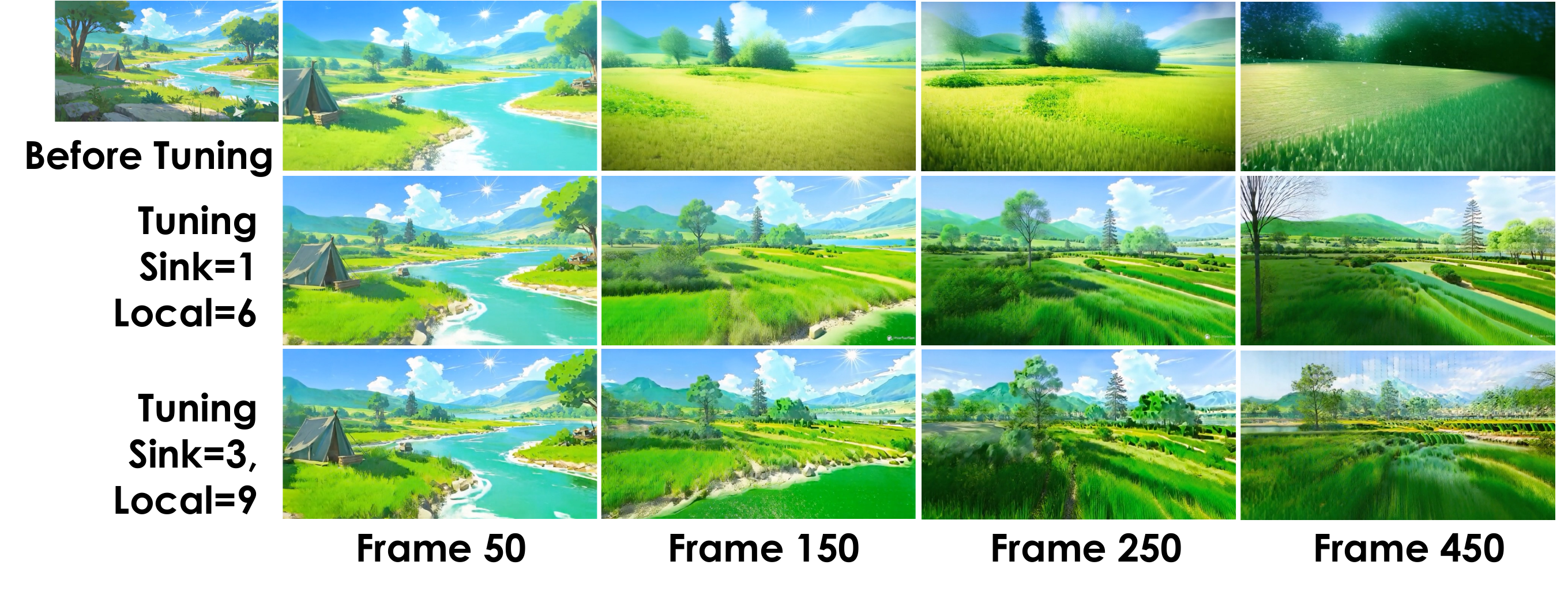}
    \vspace{-6mm}
\caption{ \textbf{Qualitative Analysis of Long-Video Tuning and Cache Settings.} \textbf{Row 1}: Baseline results without Long-Video Tuning (sink token size = 1, local attention size = 6). \textbf{Row 2}: Incorporates Long-Video Tuning upon the baseline. \textbf{Row 3}: Further modifies setting based on Row 2 by increasing the sink token size to 3 and local attention size to 9. Input prompts and camera parameters remain consistent across all samples.}
\label{fig:long_tune_ab}
\end{figure}

\paragraph{Analysis of Long-Video Tuning and Cache Settings} We qualitatively analyze the impact of long-video tuning and KV cache settings, specifically regarding the sink tokens and local attention. As illustrated in Figs.~\ref{fig:long_tune_ab} , where we compare generated frames at aligned time steps, the integration of randomized extended long-video tuning substantially improves video fidelity and motion consistency beyond the 450th frame. Moreover, expanding the sink tokens and local attention size can enrich the detail but increases the artifacts. These observations confirm both the efficacy of our tuning strategy and the importance of leveraging sink tokens and local attention to maintain robust context.

\section{Limitation and Future Work}
\label{sec:limitation}
Despite its advancements, our framework has several limitations that highlight avenues for future research. First, while our randomized long-video tuning strategy alleviates error accumulation in autoregressive generation, it does not entirely eliminate it, and semantic drift may still manifest in long sequences (More than 500 frames). This is partly attributable to our model's lack of an explicit long-term memory mechanism, a crucial component for advanced world models, as it relies instead on the finite capacity of its KV cache. Furthermore, the scope of supported interactions is currently centered on single-step, immediate-effect actions. Enabling multi-stage tasks that require logical reasoning or planning remains a significant future challenge. Finally, although we achieve real-time performance at 16 FPS, further optimization is required to reduce latency for highly reactive gameplay and to enable deployment on more accessible hardware.

\section{Conclusion}
\label{sec:conclusion}

In this work, we introduced \textbf{Hunyuan-GameCraft-2}, an interactive game world model capable of generating high-fidelity, controllable video in response to free-form text instructions and keyboard/mouse actions. We formally defined interactive video data and proposed automated pipelines for its curation and synthesis, effectively addressing the data bottleneck that has hindered progress in this domain. Our model unifies multimodal control signals within a robust training framework, leveraging a novel randomized long-video tuning scheme and efficient inference mechanisms like KV-recache to achieve stable, long-horizon, and real-time interactive generation.
To rigorously evaluate our contributions, we introduced \textbf{InterBench}, a new benchmark specifically designed to assess action-level interaction quality. Extensive experiments demonstrate that GameCraft-2 significantly outperforms existing state-of-the-art models across all dimensions of interaction fidelity, visual quality, and temporal coherence. By pushing the frontier from passive video synthesis to active, user-driven world generation, our work marks a significant step toward creating truly playable and immersive AI-generated virtual experiences.

\newpage
{
    \small
    \bibliographystyle{ieeenat_fullname}
    \bibliography{main}
}

 
\clearpage
\appendix


\section{Illustrative Examples of Interactive Video Data}

To provide a comprehensive understanding of our definition, we present representative examples that clarify the boundary between interactive and non-interactive video data. 

\subsection{Positive Examples: Interactive Video Data}

The following examples satisfy one or more properties of interactive video data, exhibiting clear causal structures and perceivable state transitions.

\paragraph{Subject Emergence.}
\begin{itemize}[leftmargin=12pt]
    \item \textit{Example 1: Vehicle Appearance.} An empty street (initial state) transitions as a car enters from off-screen and parks at the roadside (transition process), culminating in a scene depicting \textit{``a car parked on the street''} (final state). The automobile constitutes the emergent core subject, transforming the scene from vacant to occupied.
    
    \item \textit{Example 2: Object Retrieval.} From a first-person perspective, the frame initially contains only a pair of hands (initial state). The hands retrieve a key from a pocket and hold it prominently (transition process), resulting in a final state of \textit{``hands holding a key'' }(final state). The key represents the emergent core subject.
\end{itemize}

\paragraph{Action-Driven Interaction.}
\begin{itemize}[leftmargin=12pt]
    \item \textit{Example 3: Door Opening.} The scene begins with a subject standing before a closed door (initial state). The subject pushes the door open (transition process), leading to a fully open door (final state). This exemplifies direct interaction where the subject acts upon an object, inducing a clear state change.
    
    \item \textit{Example 4: Weapon Discharge.} A character aims a firearm at a target (initial state), pulls the trigger (transition process), resulting in projectile impact and target destruction (final state). This demonstrates action-consequence causality with observable physical effects.
\end{itemize}

\paragraph{Environmental State Evolution.}
\begin{itemize}[leftmargin=12pt]
    \item \textit{Example 5: Weather Transition.} A clear sky (initial state) undergoes gradual cloud accumulation followed by snowfall of increasing intensity (transition process), ultimately blanketing the entire scene in heavy snow (final state). This represents a fundamental transformation of the environmental weather attribute.
    
    \item \textit{Example 6: Spatial Transition.} Upon opening a door, the camera view shifts from an interior room (initial scene) to an exterior courtyard (final scene). This exemplifies a discrete scene transition driven by subject action, fundamentally altering the observational context.
\end{itemize}

\subsection{Negative Examples: Non-Interactive Video Data}

These examples, though visually dynamic, lack the defining characteristics of interactive video data.

\paragraph{Continuous Static Process.}
\begin{itemize}[leftmargin=12pt]
    \item \textit{Example 1: Sustained Blizzard.} A 10-second video segment depicting continuous heavy snowfall. Although visually dynamic, the macroscopic state remains constant as \textit{``actively snowing throughout,''} lacking a transition from \textit{``no snow''} to \textit{``snow present.''} The absence of state evolution disqualifies this as interactive data.
\end{itemize}

\paragraph{Stochastic Background Activity.}
\begin{itemize}[leftmargin=12pt]
    \item \textit{Example 2: Busy Intersection.} A scene featuring continuous pedestrian and vehicular traffic at a crowded intersection. While abundant motion exists, there is no singular event-driven macroscopic state change with definitive beginning and end points. The scene's overarching state persistently remains \textit{``busy intersection,''} lacking a coherent causal narrative.
\end{itemize}

\paragraph{Generalized Motion without Core Subject.}
\begin{itemize}[leftmargin=12pt]
    \item \textit{Example 3: Ambient Environmental Fluctuations.} Ripples propagating across a water surface or leaves swaying in wind. These phenomena typically constitute random environmental perturbations rather than state transitions driven by specific subjects or events with explicit causal chains. They lack the purposeful, agent-driven transformation characteristic of interactive data.
\end{itemize}

\subsection{Interaction Categories}
\label{sec:appendix_categories} 

Following the definition of interactive data in the main text, we provide here a detailed breakdown of the three principal interaction categories used to structure our dataset and analysis. Each category includes both simple and complex settings to reflect different levels of difficulty and to facilitate a fine-grained evaluation of model capabilities.

\paragraph{(1) Environmental Interactions.}  
These interactions reflect global or local scene changes.  
\textit{Simple cases} include atmospheric effects such as \textit{snowfall} and \textit{rainfall}.  
\textit{Complex cases} involve more substantial causal transformations, such as \textit{lightning strikes} or \textit{triggering an explosion}, which require coherent illumination changes, particle dynamics, and physically plausible propagation.

\paragraph{(2) Actor Actions.}  
These interactions are driven by an embodied or first-person actor.  
\textit{Simple cases} include basic manipulation actions such as \textit{drawing a gun} or \textit{drawing a knife}.  
\textit{Complex cases} require multi-step or environment-affecting interactions, such as \textit{drawing a torch to illuminate the surroundings}, \textit{firing a gun}, \textit{taking out a phone and operating it}, or \textit{opening a door}. These demand consistent body–object coordination and temporal stability.

\paragraph{(3) Entity and Object Appearances.}
These interactions introduce new entities into the scene.
\textit{Simple cases} include the appearance of a single human or common object.
\textit{Complex cases} involve entities with more distinct geometry or motion priors, such as animals (cat, dog, deer, wolf, dragon) or vehicles (red SUV, yellow sports car, blue truck, black off-road vehicle), which require accurate spatial placement, scale consistency, and stable identity preservation.


\section{Dataset Showcase}
\label{sec:appendix_dataset_showcase}

This appendix provides visual examples from our constructed dataset, which is composed of two primary sources: curated real-world gameplay footage and synthetically generated interactive videos. The following sections showcase the diversity and quality of each data type.

\subsection{Curated Gameplay Data}
\begin{figure*}[hbt!]
    \centering
    \includegraphics[width=\linewidth]{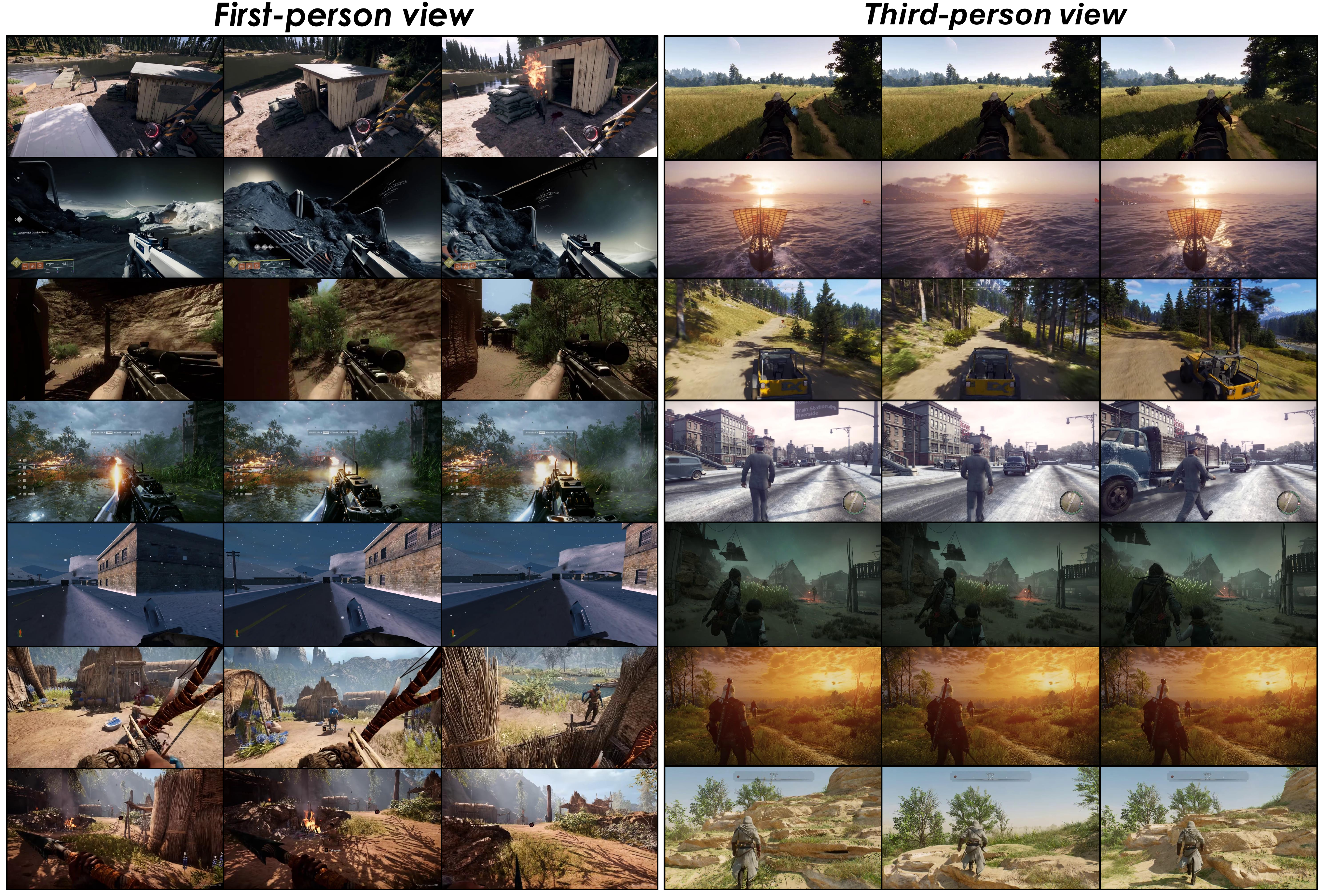}
\caption{\textbf{Examples of First-person and Third-person Interactive Gameplay Videos.}
Samples showing diverse actor actions under different viewpoints, illustrating rich interactive semantics captured from our gameplay collection.}
    \label{fig:curation_data_showcase}
\end{figure*}

\begin{figure*}[hbt!]
    \centering
    \includegraphics[width=\linewidth]{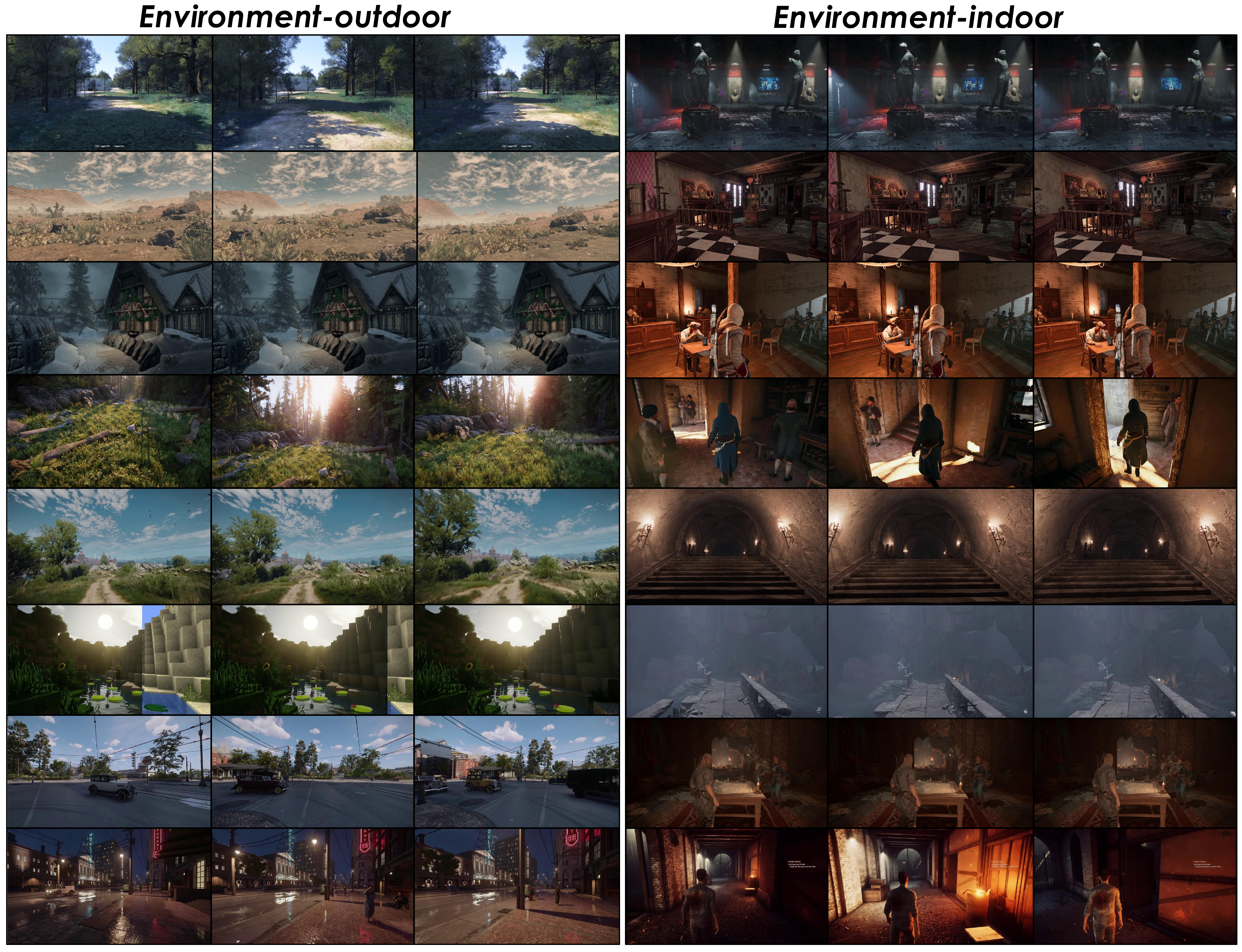}
\caption{\textbf{Comprehensive Environment Diversity in Our Dataset.}
Examples of outdoor and indoor environments across natural and urban scenes, under diverse lighting, weather, and terrain conditions.}
\label{fig:curation_data_showcase_2}
\end{figure*}

The following figures illustrate the rich diversity of our curated gameplay data, collected from over 150 AAA games. As shown, the dataset covers a wide array of interaction contexts, including both first-person and third-person viewpoints (Fig.~\ref{fig:curation_data_showcase}), as well as a comprehensive range of environments spanning natural and urban scenes under various lighting, weather, and terrain conditions (Fig.~\ref{fig:curation_data_showcase_2}). This diversity is crucial for training robust and generalizable world models.

\subsection{Synthetic Interaction Data}
\begin{figure*}[h!]
    \centering
    \includegraphics[width=\linewidth]{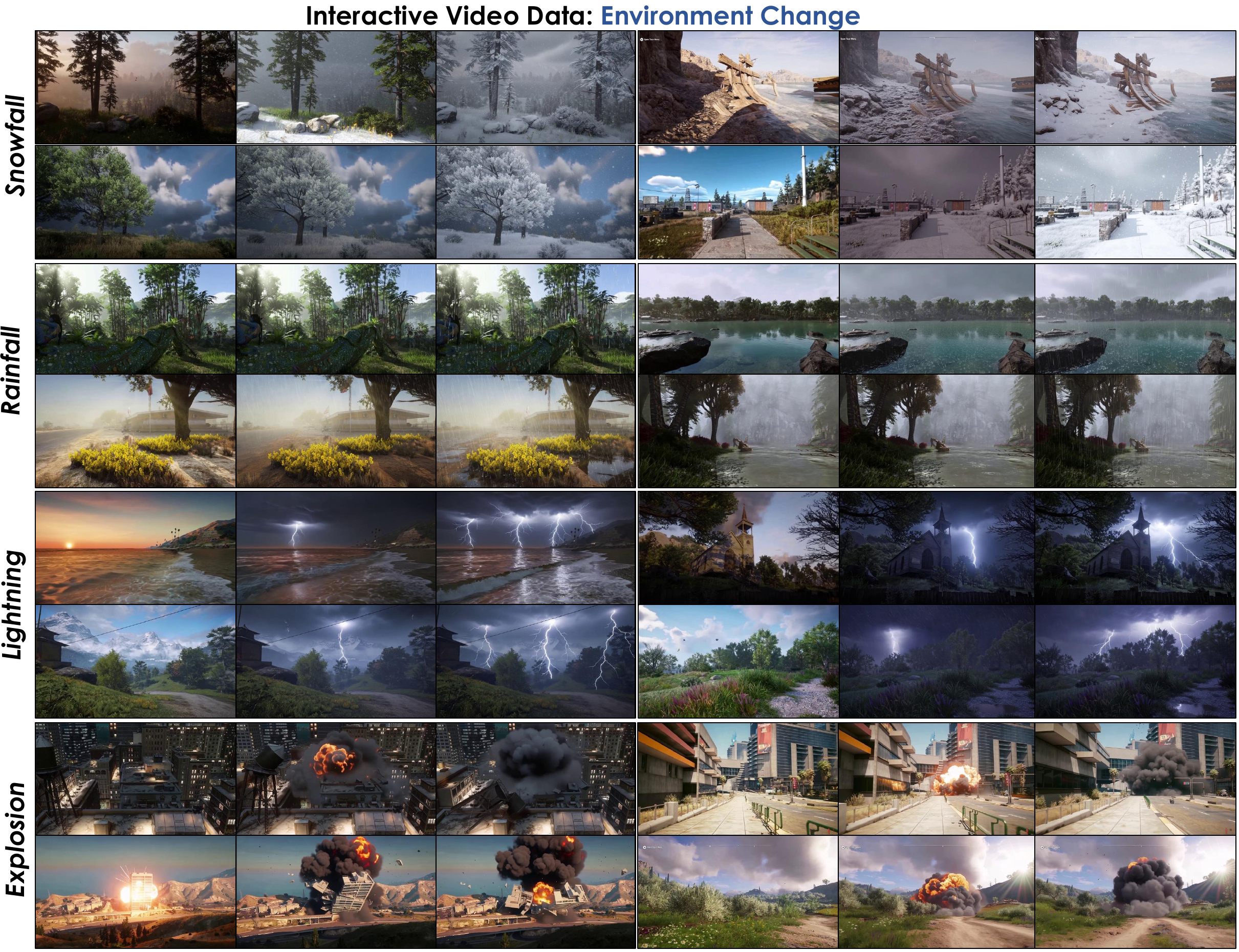}
\caption{\textbf{Synthetic Examples of Environmental Interactions.}
Examples of synthetic scene-change interactions generated by our pipeline, covering \textit{snowfall}, \textit{rainfall}, \textit{lightning}, and \textit{explosions}.}
    \label{fig:Construction_data_enirvonment}
\end{figure*}

\begin{figure*}[h!]
    \centering
    \includegraphics[width=\linewidth]{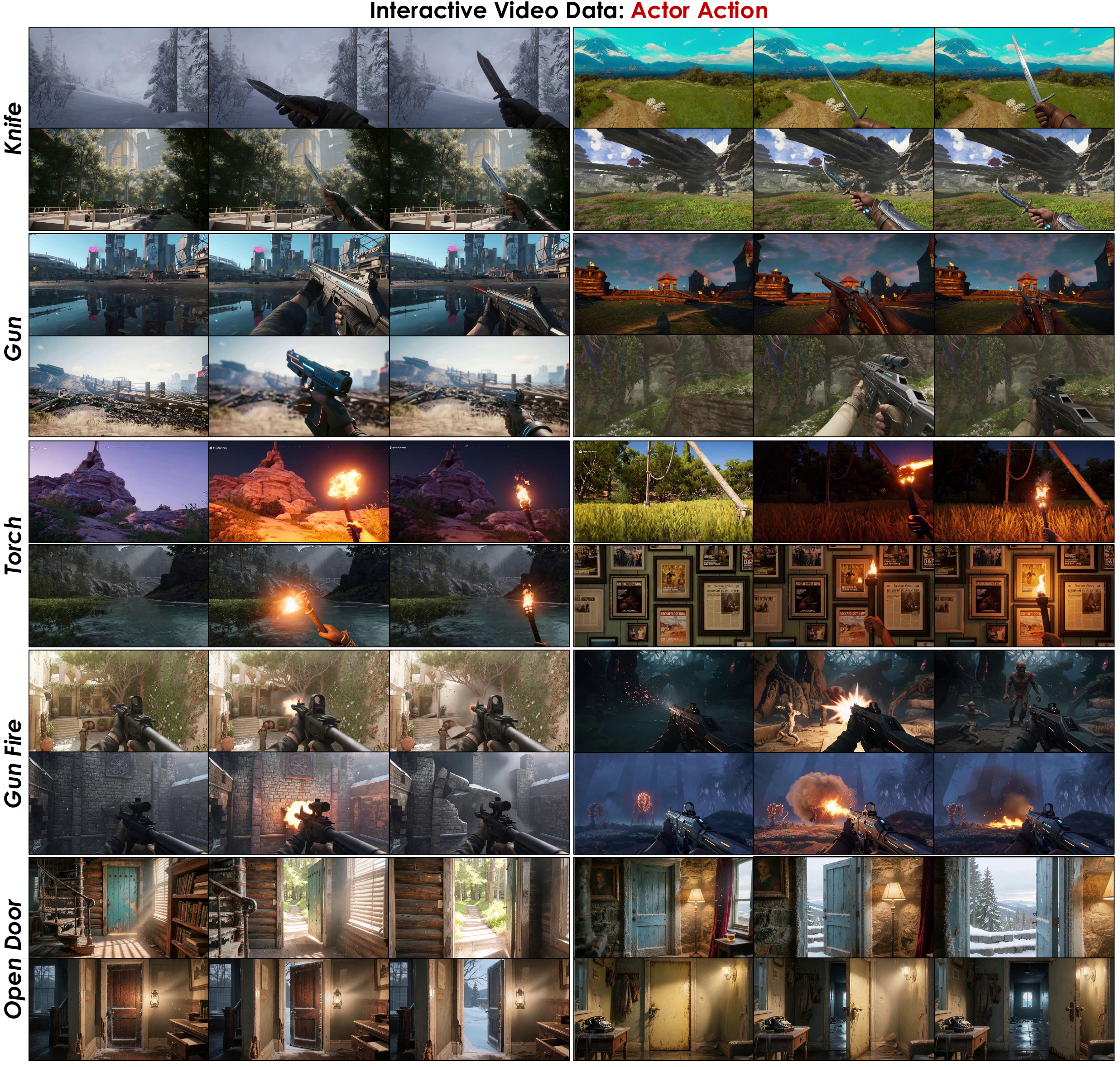}
    \vspace{-3mm}
\caption{\textbf{Synthetic Examples of Actor Actions.}
Examples illustrating the range of interactive behaviors synthesized by our pipeline. 
\textbf{\textit{Actor-driven interactions}} include \textit{drawing a knife}, \textit{drawing a gun}, 
\textit{drawing a torch}, \textit{firing a weapon}, and \textit{opening a door}, 
which require consistent body–object coordination and temporal coherence. }
\label{fig:Construction_data_actor_action}
\end{figure*}

\begin{figure*}[h!]
    \centering
    \includegraphics[width=\linewidth]{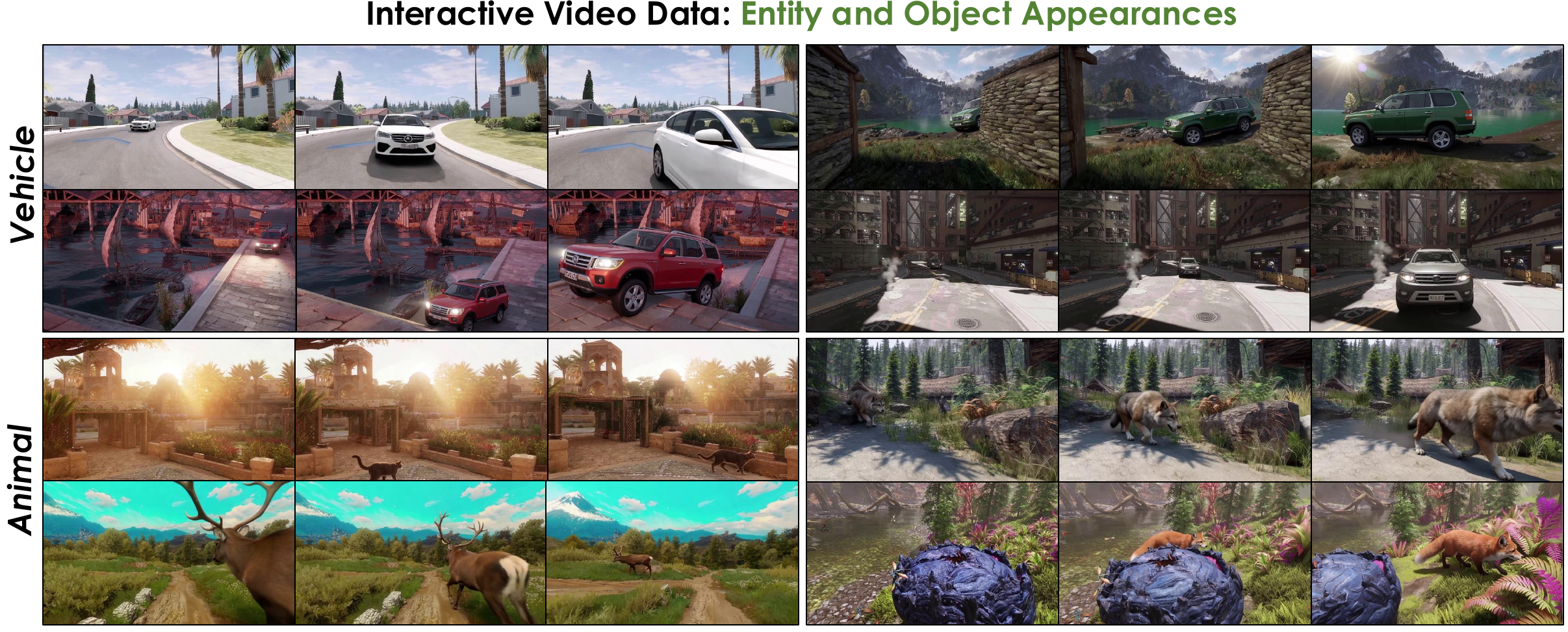}
    \vspace{-3mm}
\caption{\textbf{Synthetic Examples of Entity/Object Appearances.}
Examples illustrating the range of interactive behaviors synthesized by our pipeline. 
\textbf{\textit{Entity- and object-level interactions}} include \textit{animal intrusion} and \textit{vehicle entry}, 
showing the pipeline’s capability to introduce new entities with realistic geometry, scale consistency, and stable identity across frames.}
\label{fig:Construction_data_entity_appearance}
\end{figure*}

Generated by our synthetic data pipeline, the following examples demonstrate the pipeline's capability to create controlled and high-quality interactive videos. These examples cover the three main interaction categories defined in our work: \textbf{Environmental Interactions} such as weather changes and explosions (Fig.~\ref{fig:Construction_data_enirvonment}), \textbf{Actor Actions} involving complex body-object coordination (Fig.~\ref{fig:Construction_data_actor_action}), and \textbf{Entity/Object Appearances} that introduce new subjects into the scene with high fidelity (Fig.~\ref{fig:Construction_data_entity_appearance}).

\section{Detailed Comparison Across Interaction Dimensions}
\begin{figure*}[h!]
    \centering
    \includegraphics[width=\linewidth]{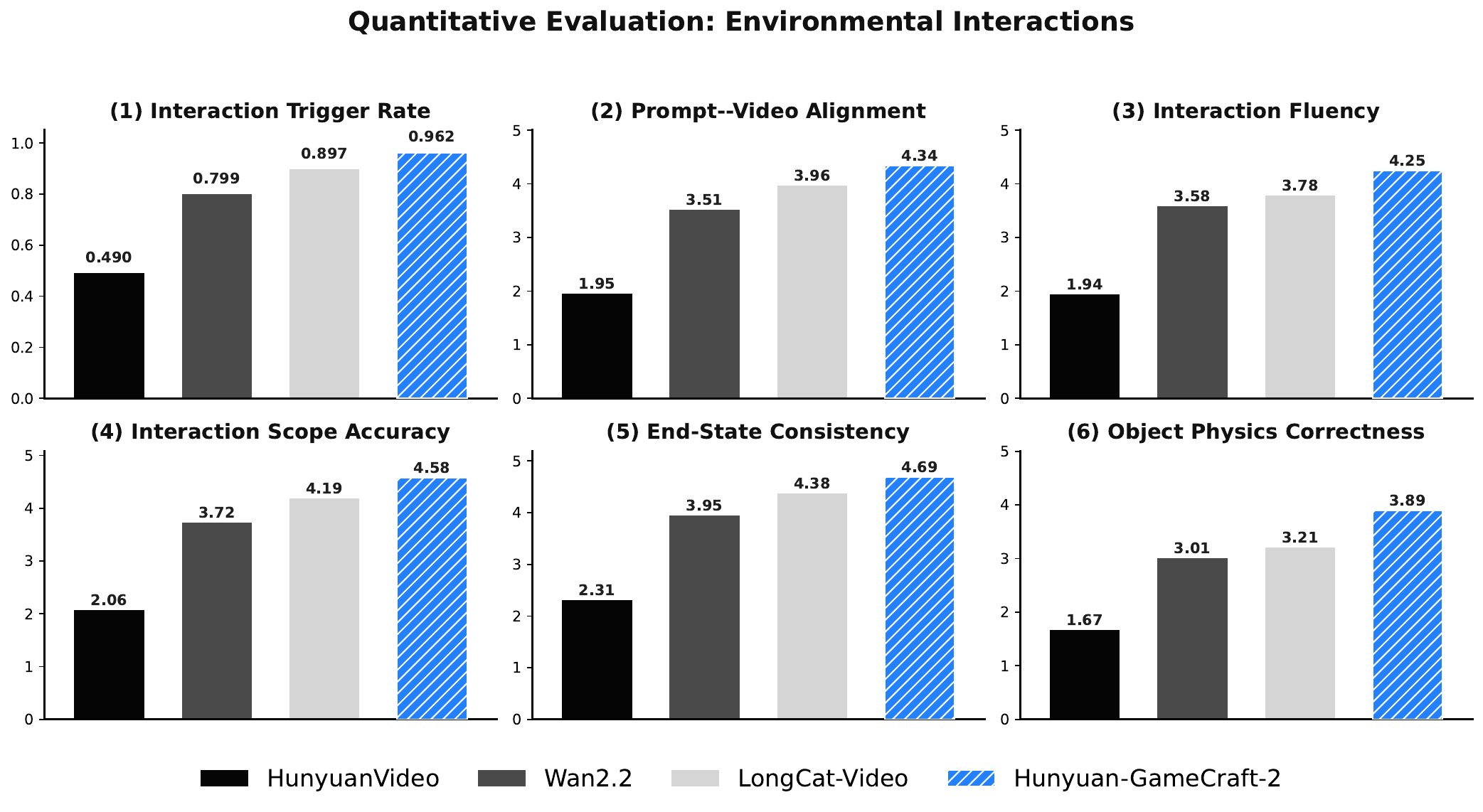}
    \caption{\textbf{Comparison of Environmental Interactions with Baseline Models.} 
Qualitative results showing the fidelity and consistency of environment-level effects. Our approach better preserves global influence and temporal stability.}
    \label{fig:qual_env}
\end{figure*}

\begin{figure*}[h!]
    \centering
    \includegraphics[width=\linewidth]{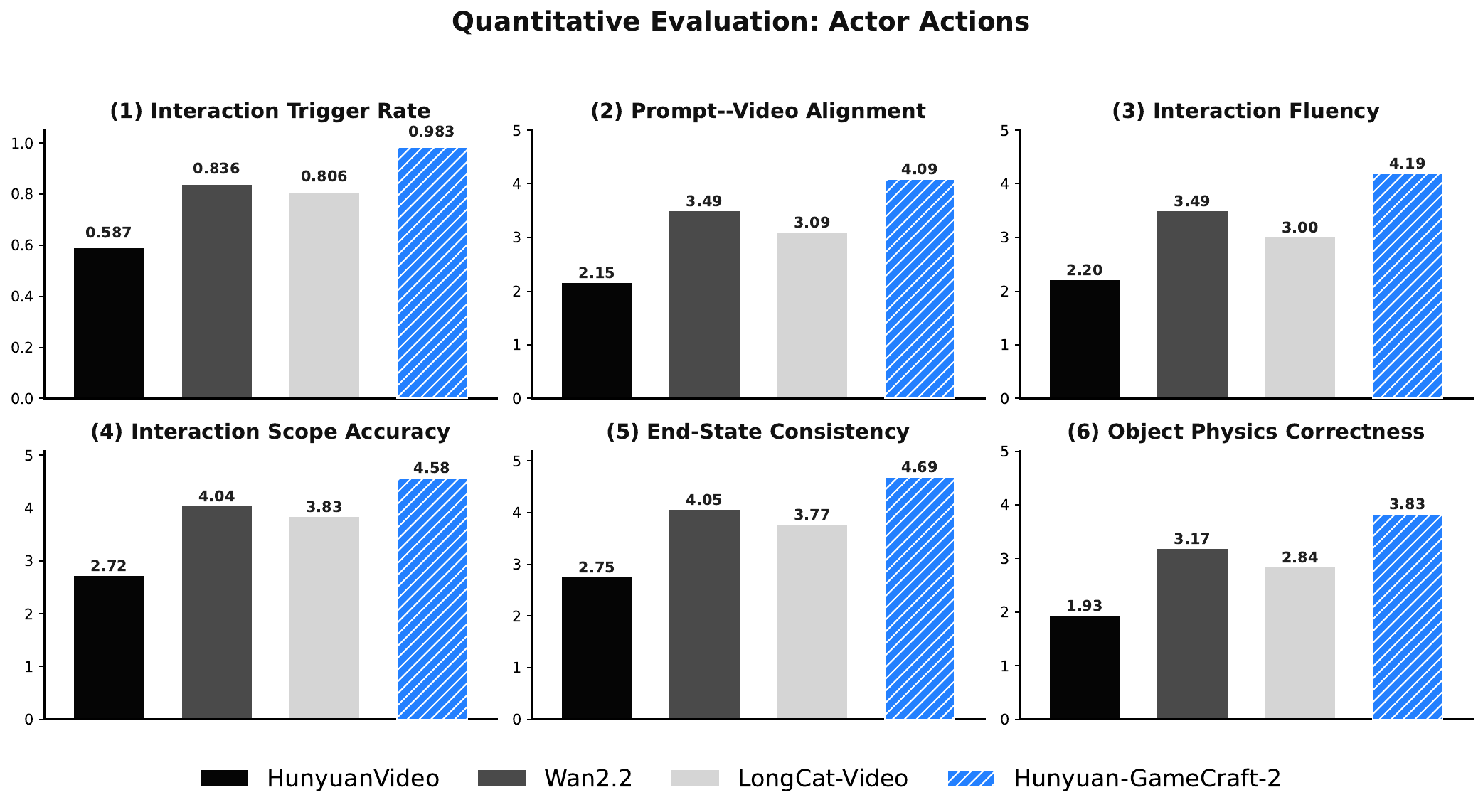}
    \caption{\textbf{Comparison of Actor-Action Interactions with Baseline Models.} 
Visual comparisons illustrating the quality of action-level interactions across representative prompts. Our method produces more coherent and physically consistent actions than all baselines.}
    \label{fig:qual_action}
\end{figure*}

\begin{figure*}[h!]
    \centering
    \includegraphics[width=\linewidth]{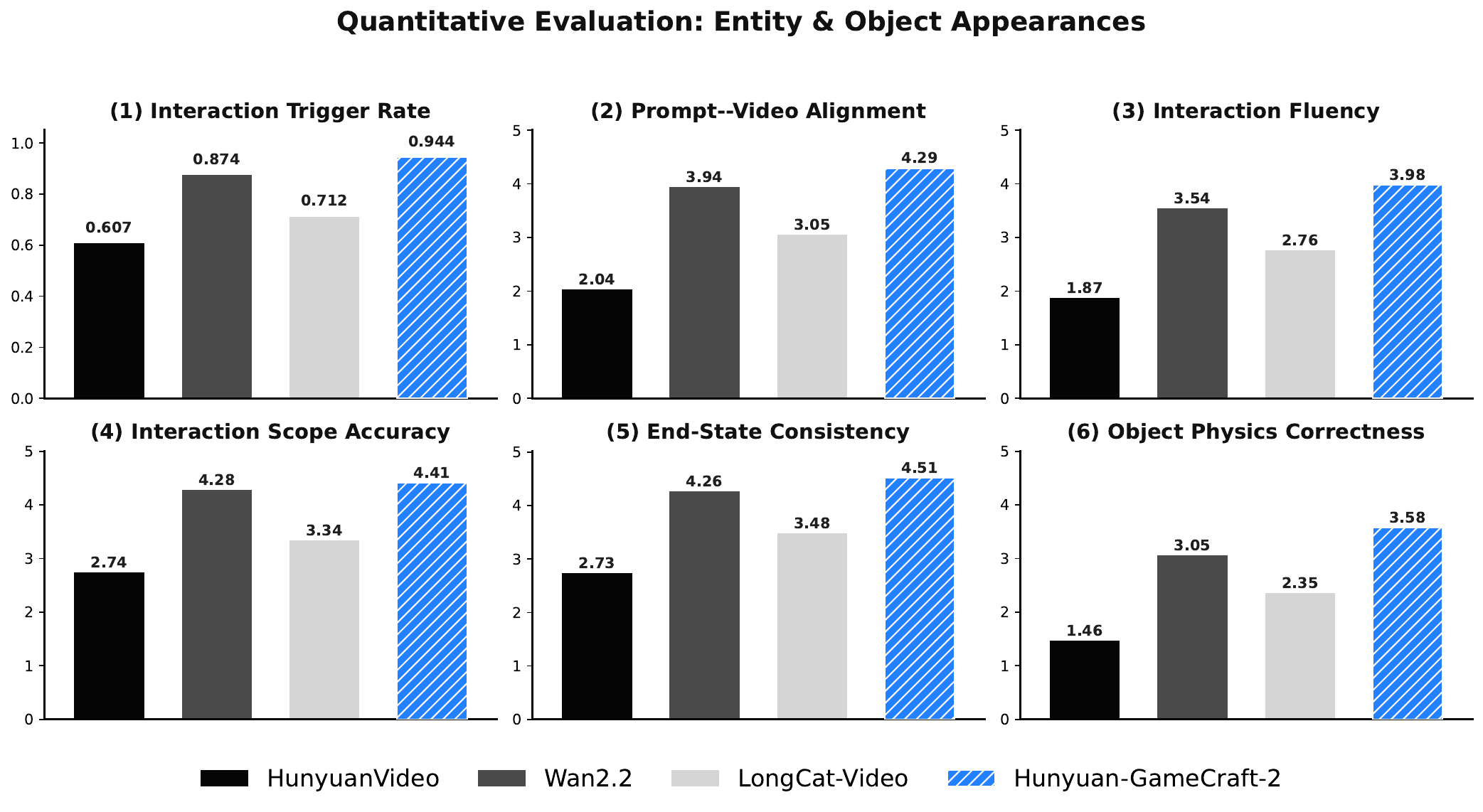}
    \caption{\textbf{Comparison of Entity and Object Appearance Interactions with Baseline Models.} 
Visual comparisons of object emergence and interaction correctness. Our method delivers more accurate, stable, and physically plausible object behaviors.}
    \label{fig:qual_entity}
\end{figure*}

In this section, we present a concise comparison of our approach against baseline models across three key dimensions of interactive video generation: environmental interactions, actor–action dynamics, and entity or object appearance behaviors. As illustrated in Figures~\ref{fig:qual_env}--\ref{fig:qual_entity}, our method achieves higher temporal stability, more coherent action execution, and more accurate object emergence, consistently outperforming existing models across diverse interaction scenarios.

\clearpage

\section{InterBench: A Detailed Protocol for Benchmarking Action-Level Interaction}
\label{sec:appendix_interbench}

\paragraph{Motivation and Design Philosophy.}
Existing video generation benchmarks, such as Fréchet Video Distance or CLIP Score, primarily assess perceptual quality, temporal consistency, and static text-video alignment. While valuable, they are ill-suited for evaluating \textit{interactive} video generation, where the primary task is to render a causal change in response to a specific action command. These metrics cannot distinguish between a correctly executed action and a visually plausible but semantically incorrect video. To fill this critical gap, we designed \textbf{InterBench}, an evaluation protocol specifically tailored to measure the fidelity of action-level interactions. Its philosophy is to deconstruct the complex concept of a ``good interaction'' into a set of distinct, measurable, and interpretable dimensions, enabling a fine-grained analysis of model capabilities and failure modes.

\paragraph{Interaction Trigger Rate.}
This dimension serves as the most fundamental, gateway assessment. It asks the question: \textit{Did the requested interaction happen at all?''} This metric is designed to isolate the model's basic ability to acknowledge and act upon an instruction, separating cases where the model successfully initiated the action from those where it completely failed to respond. This is a binary metric: 
\begin{itemize} \item \textbf{1 (Success):} The requested interaction is initiated in the video. For instance, for the prompt \textit{draw a gun,''} this score is given if a gun becomes visible. If this score is given, the subsequent dimensions are evaluated on their respective scales.
\item \textbf{0 (Failure):} The requested interaction does not occur at all. The model ignores or completely misunderstands the interaction prompt. If this score is given, all subsequent dimensions are automatically scored 0.
\end{itemize}

\paragraph{Prompt--Video Alignment.}
Beyond simply triggering an action, this dimension evaluates the semantic fidelity of the generated video with respect to the \textit{entire} prompt (both the base scene description and the interaction command). It ensures the interaction happens in the \textit{right way} and the \textit{right context}, encompassing both static and dynamic alignment.
This metric is scored on a 0-1-3-5 ordinal scale, contingent on the interaction being triggered:
\begin{itemize}[leftmargin=*]
    \item \textbf{5 (Excellent):} Both the static context (scene, style) and the dynamic action perfectly match the prompt's description.
    \item \textbf{3 (Moderate):} The primary action is correct, but there are minor semantic deviations in the scene's context or the specifics of the action's execution.
    \item \textbf{1 (Poor):} A recognizable interaction occurs, but it involves a major semantic error, such as performing the wrong action (e.g., closing instead of opening a door) or generating a scene that bears no resemblance to the base prompt.
    \item \textbf{0 (Failure):} The triggered video content shows no meaningful semantic alignment with either the prompt's context or its specified action.
\end{itemize}

\paragraph{Interaction Fluency.}
This dimension measures the temporal naturalness and continuity of the interaction process. It specifically penalizes temporal discontinuities such as abrupt teleportation of objects, noticeable frame jumps, unrealistic motion jitter, and structural tearing of geometry, particularly around the interacting regions.
This metric is scored on a 0-1-3-5 ordinal scale:
\begin{itemize}[leftmargin=*]
    \item \textbf{5 (Excellent):} The motion is perfectly smooth, continuous, and natural, with no temporal artifacts present.
    \item \textbf{3 (Moderate):} The motion is generally continuous but contains minor, non-disruptive artifacts like slight jitter or a single inconspicuous jump-cut.
    \item \textbf{1 (Poor):} The interaction is plagued by severe temporal artifacts (e.g., constant flickering, object teleportation) that significantly disrupt the viewing experience.
\end{itemize}

\paragraph{Interaction Scope Accuracy.}
This metric assesses a model's spatial reasoning by examining whether the spatial extent and environmental influence of an interaction are plausible and consistent with its expected scope (global or local).
This metric is scored on a 0-1-3-5 ordinal scale:
\begin{itemize}[leftmargin=*]
    \item \textbf{5 (Excellent):} The spatial influence of the interaction is physically and semantically correct (e.g., global effects are global, local effects are local and propagate realistically).
    \item \textbf{3 (Moderate):} The scope is generally correct but with minor inaccuracies, such as a global effect not covering the entire scene or a local effect having a slightly incorrect area of influence.
    \item \textbf{1 (Poor):} The scope is fundamentally wrong. For example, a global event is rendered as a tiny local patch, or a local effect implausibly affects the entire scene.
\end{itemize}

\paragraph{End-State Consistency.}
A successful interaction must not only be initiated correctly but also \textit{converge} to a stable and correct outcome. This dimension evaluates the final state of the video to ensure the result of the action persists as expected.
This metric is scored on a 0-1-3-5 ordinal scale:
\begin{itemize}[leftmargin=*]
    \item \textbf{5 (Excellent):} The interaction converges to the correct final state, which remains stable until the end of the video.
    \item \textbf{3 (Moderate):} The final state is mostly correct but exhibits minor instability, such as slight flickering, object drift, or subtle geometric inconsistencies.
    \item \textbf{1 (Poor):} The interaction fails to converge correctly. The final state is incorrect, highly unstable (e.g., oscillating), or the effects of the action vanish prematurely.

\end{itemize}

\paragraph{Object Physics Correctness.}
This dimension focuses on the physical plausibility and structural integrity of the objects and agents involved in the interaction, evaluating whether their behavior adheres to basic physical principles like object permanence, rigidity, and kinematics.
This metric is scored on a 0-1-3-5 ordinal scale:
\begin{itemize}[leftmargin=*]
    \item \textbf{5 (Excellent):} All objects and agents maintain structural integrity and interact in a physically plausible manner. There is no unnatural deformation, interpenetration, or kinematic errors.
    \item \textbf{3 (Moderate):} Minor physical inaccuracies are present, such as slight object warping during movement or brief, non-critical interpenetration between an agent and an object.
    \item \textbf{1 (Poor):} Severe physical violations occur. Objects unnaturally deform, agents pass through solid objects, or motion is kinematically impossible.
\end{itemize}

\end{document}